%% file: neurips_2025.tex
\documentclass{article}

\PassOptionsToPackage{numbers, compress}{natbib}

  \usepackage[preprint]{neurips_2025}

\usepackage[utf8]{inputenc} %
\usepackage[T1]{fontenc}    %
\usepackage{hyperref}       %
\usepackage{url}            %
\usepackage{booktabs}       %
\usepackage{amsfonts}       %
\usepackage{nicefrac}       %
\usepackage{microtype}      %
\usepackage{xcolor}         %

\usepackage{acronym}
\usepackage{graphicx}
\usepackage{multirow}
\usepackage{makecell}
\usepackage{longtable}
\usepackage{subcaption}
\usepackage{caption}
\usepackage[flushleft]{threeparttable}
\usepackage{rotating}

\title{AquaMonitor: A multimodal multi-view image sequence dataset for real-life aquatic invertebrate biodiversity monitoring}

\author{%
    Mikko Impiö$^1$,
    Philipp M. Rehsen$^{2,3}$,
    Tiina Laamanen$^1$, \\
    \textbf{Arne J. Beermann}$^{2,3}$,
    \textbf{Florian Leese}$^{2,3}$,
    \textbf{Jenni Raitoharju}$^{4,1}$ \\
 $^1$ Finnish Environment Institute, Finland, \\
 $^2$ Aquatic Ecosystem Research, University of Duisburg-Essen, Germany, \\
 $^3$ Centre for Water and Environmental Research (ZWU), \\
 University of Duisburg-Essen, Germany, \\
 $^4$ Faculty of Information Technology, University of Jyväskylä, Finland \\
\small{\url{https://huggingface.co/datasets/mikkoim/aquamonitor}}
\\
\small{\url{https://github.com/mikkoim/aquamonitor}}
}

\begin{document}

\newacro{WFD}[WFD]{Water Framework Directive}

\maketitle

\vspace{-6mm}
\begin{abstract}

This paper presents the AquaMonitor dataset, the first large computer vision dataset of aquatic invertebrates collected during routine environmental monitoring. While several large species identification datasets exist, they are rarely collected using standardized collection protocols, and none focus on aquatic invertebrates, which are particularly laborious to collect. For AquaMonitor, we imaged all specimens from two years of monitoring whenever imaging was possible given practical limitations. The dataset enables the evaluation of automated identification methods for real-life monitoring purposes using a realistically challenging and unbiased setup. The dataset has 2.7M images from 43,189 specimens, DNA sequences for 1358 specimens, and dry mass and size measurements for 1494 specimens, making it also one of the largest biological multi-view and multimodal datasets to date. We define three benchmark tasks and provide strong baselines for these: 1) Monitoring benchmark, reflecting real-life deployment challenges such as open-set recognition, distribution shift, and extreme class imbalance, 2) Classification benchmark, which follows a standard fine-grained visual categorization setup, and 3) Few-shot benchmark, which targets classes with only few training examples from very fine-grained categories. Advancements on the Monitoring benchmark can directly translate to improvement of aquatic biodiversity monitoring, which is an important component of regular legislative water quality assessment in many countries.
\end{abstract}

\captionsetup[figure]{font=small}
\captionsetup[table]{font=small}

\vspace{-4mm}
\input{1_intro}

\input{2_related}

\input{3_dataset}
\input{4_experiments}

\input{5_conclusions}

\newpage

\bibliographystyle{abbrvnat}
\bibliography{references.bib}

\appendix

\input{X_suppl}

\clearpage
\newpage

\end{document}

%% file: 1_intro.tex
\section{Introduction}
\label{sec:intro}

Computer vision has been recognized as an important technology for next-generation biodiversity monitoring, enabling monitoring and collection of environmental information on a global scale \cite{hoye2021Deep, oliver2023Camera, tuia2022Perspectives, vanklink2022Emerging, besson2022fully, vanklink2024toolkit, gonzalez2023global, gonzalez2023framework, wagner2020Insect}.
Advancements in computer vision methods, such as fine-grained visual categorization, few-shot learning, domain adaptation, and out-of-distribution detection, have applications in biodiversity monitoring and contribute to the progress of new methods in the field \cite{vanhorn2021Benchmarking, lu2022Fewshot, koh2020WILDS}. However, popular benchmarks and datasets overrepresent charismatic species, such as mammals and birds \cite{beery2021iWildCam, WahCUB_200_2011, vanhorn2015Building}, while groups with a high need of monitoring, such as insects and other invertebrates, have gained less attention \cite{hoye2021Deep, schneider2023Getting, pawar2003Taxonomic}.

Efforts are being made to improve automated identification of insects and other invertebrates, due to their importance as providers of ecosystem services \cite{noriega2018Research, seibold2019Arthropod} and alarming decline \cite{hallmann2017More, wagner2020Insect, vanklink2024Disproportionate}. Especially aquatic species serve as important indicators of water quality, and legislation in many countries, e.g., the EU Water Framework Directive \cite{european2000directive}, mandates their monitoring.
Recent studies have proposed innovative monitoring methods \cite{hoye2021Deep, vanklink2024toolkit}, imaging devices \cite{arje2020Automatic, schneider2022Bulk, wuhrl2022DiversityScanner, deschaetzen2023Riverine}, and datasets \cite{gharaee2024BIOSCAN5M, jain2024Insect, stevens2024BioCLIP, maruf2024VLM4Bio} that contribute to solving problems related to data acquisition and processing of biodiversity information.
The field is gaining technological maturity in the sense that good performance has been demonstrated on datasets with closed-set categories \cite{besson2022fully, schneider2023Getting, arje2020Automatic, hoye2022Accurate}.

The ecological and computer vision communities are still lacking image datasets from realistic biodiversity monitoring setups for invertebrates.
Related deep learning datasets are commonly from toy problems or do not represent a true distribution of data and taxa (e.g., species or genera).
Selection bias becomes an issue, when categories are chosen based on their availability or fine-grained labels being merged to broader groups to ease identification.
In particular, the rarest taxa are frequently ignored due to the lack of sufficient training data.
However, such taxa should not be left out when evaluating the suitability of computer vision methods for practical monitoring.
Similarly, the geographical and temporal information is commonly not provided.

\noindent Our main contributions can be summarized as follows: 
\begin{itemize}
\item	We present the AquaMonitor dataset, a novel collection of 44,854 multi-view image sequences (2.7M images) of aquatic invertebrates, representing 43,189 specimens from an routine freshwater monitoring program \cite{vilmi2021maa} over two years.
\item The AquaMonitor dataset contains useful information typically missing from existing computer vision datasets, including sampling site and sampling time for each specimen. We also include additional modalities: individual DNA sequences, biomass, and size measured for subsets of data.
\item We define three different benchmark taxonomic identification tasks and provide baseline results and pretrained models for all of them. The benchmarks are for 1) A real-life monitoring task, including all challenges naturally encountered in monitoring, including temporal dimension (training and test data collected in different years) and very long-tailed distribution with partially non-overlapping categories (i.e., out-of-distribution samples in the test set), 2) A traditional fine-grained classification task for categories with more than 50 examples, mainly intended for demonstrating that this kind of setup typically used in the currently available datasets is not suitable for evaluating methods for real-life monitoring purposes, and 3) A few-shot learning task for categories with less than 50 examples, enabling focusing on the important need for species identification systems to be able to learn new categories with just a few examples.
\item We train several strong baseline models for all the benchmarks and a biomass estimation task. We report a wide selection of comparative results and make all model weights and training codes publicly available.
\end{itemize}

%% file: 2_related.tex
\section{Related work}

Datasets from the natural world, such as Flowers102 \cite{nilsback2008Automated}, Caltech-UCSD Birds-200-2011 \cite{WahCUB_200_2011}, NABirds \cite{vanhorn2015Building}, and iNat21 \cite{vanhorn2021Benchmarking}, have proven to be popular in benchmarking various computer vision tasks, such as fine-grained \cite{wei2022FineGrained} and ultra-fine-grained \cite{yu2021Benchmark} visual categorization, few-shot learning \cite{wang2020Generalizing, rodriguez2024Recognition}, multimodal classification \cite{diao2022MetaFormer, chu2019GeoAware}, and open-set/out-of-distribution recognition \cite{geng2021Recent, scheirer2013Open, lang2024Coarse, liu2018Open, vaze2022Generalized}.
Over the past fifteen years, a lot of small computer vision datasets for insect and invertebrate identification have been also collected, usually for research purposes targeting applications in biodiversity monitoring or digitizing museum samples \cite{hansen2020Specieslevel, marques2018Ant}.
A comprehensive review of image datasets collected before 2017 is available in \cite{martineau2017survey}. More recent datasets for pest detection applications are studied in \cite{nguyen2024InsectFoundation}. However, while these small datasets can be useful for specific, niche goals, they have limitations for general biodiversity monitoring purposes \cite{schneider2023Getting}.
Data for research datasets are often chosen for practical reasons and might be biased toward taxa that are easy to identify without extensive taxonomic expertise, limiting generalizability to real-world settings.

\begin{table*}[ht]
\centering
\caption{Overview of existing large publicly available datasets containing images of invertebrates.}
\label{tab:datasets}
\resizebox{\textwidth}{!}{%
\begin{threeparttable}
\begin{tabular}{@{}llllllllrrr@{}}
\toprule
Name                                        & Year & Source &Type & Objects & Modality & Multi-view           & Taxa                         & Images & Specimens & Classes \\ \midrule
AquaMonitor $^{\dagger \ddag}$                            & 2025 &Self-imaged                     & Lab                      & Single                      & Sequence                        & \multicolumn{1}{c}{\checkmark} & Aquatic & 2.7M                       & 43,189                          & 152                           \\ \midrule
TreeOfLife-10M \cite{stevens2024BioCLIP}    & 2024 & \cite{gharaee2024step, vanhorn2021Benchmarking}, Web                    & Lab, Field               & Single                      & Image               &       &       All                                            & 10.4M                      & 10.4M                           & 454,103
\vspace{1pt}\\
\multirow{2}{*}{AMI \cite{jain2024Insect}}                   & \multirow{2}{*}{2024} & Web                    & Field           & \multirow{2}{*}{Multi}                       & \multirow{2}{*}{Image}                        &       &  \multirow{2}{*}{Flying}                                           & 2.5M                       & 2.5M                            & 5364                         \\ \vspace{1pt} 
& & Self-imaged & Camera trap & & & & & 2893 & 14,105$^\star$ & 903 \\
BIOSCAN-5M  \cite{gharaee2024BIOSCAN5M} $^\dagger$         & 2024  & Self-imaged                   & Lab                      & Single                      & Image                        &                       & Terrestrial          & 5.1M                       & 5.1M                            & 324,411                        \\
Simović et al. \cite{simovic2024Automated} & 2024 & Self-imaged & Lab & Single & Image & \multicolumn{1}{c}{\checkmark} & Aquatic & 16,650 & 5500 & 90 \\
ALUS \cite{schneider2022Bulk}              & 2022     & Self-imaged                & Lab                      & Multi                       & Image                        &                       & Flying   & 516                     & 13,059                          & 20                          \\
Høye et al. \cite{hoye2022Accurate}         & 2022  & Self-imaged                   & Lab                      & Single                      & Sequence                        & \multicolumn{1}{c}{\checkmark} &        Aquatic                    & 148,228                    & 1120                            & 16                            \\
iNaturalist  \cite{vanhorn2021Benchmarking, vanhorn2018INaturalist} & 2021  & Citizen science                   & Field                    & Single                      & Image                        &                       &                       All       & 3.2M                       & 3.2M                            & 10,000                        \\
FINBenthic2 \cite{arje2020Human}            & 2020  & Self-imaged                   & Lab                      & Single                      & Sequence                        & \multicolumn{1}{c}{\checkmark} &        Aquatic                       & 460,009                    & 9631                            & 39                            \\
Hansen et al. \cite{hansen2020Specieslevel} & 2020   & Self-imaged                  & Lab                      & Multi                       & Image                        &                       &            Beetles                 & 63,364                     & 63,364                          & 291                           \\
IP102 \cite{wu2019IP102}                    & 2019 & Web                    & Field                    & Single                      & Image                        &                       &                  Pests            & 75,222                     & 75,222                          & 102                           \\
AntNet \cite{marques2018Ant}                & 2018  & Self-imaged                   & Lab                      & Single                      & Image                        & \multicolumn{1}{c}{\checkmark} &          Ants                    & 150,088                    & 44,806                          & 57\\
\bottomrule
\end{tabular}
\begin{tablenotes}
\footnotesize
\item $\star$ 52,948 including unidentified specimens, $\dagger$ Has DNA metadata, $\ddag$ Has specimen-level biomass
\end{tablenotes}
\end{threeparttable}}
\vspace{-4mm}
\end{table*}

Existing datasets can be roughly divided into four groups based on the imaging environment (lab/field) and number of objects present in the images (single/multiple) \cite{martineau2017survey, schneider2022Bulk}.
Online image repositories, such as GBIF / iNaturalist \cite{gbif2024GBIF}, BOLD \cite{ratnasingham2007BOLD}, and BugNet, \cite{BugNet}, usually contain images of single specimens taken in various settings and often captured by citizen scientists. There are large deep learning-ready datasets collected from these sites, such as the iNaturalist datasets \cite{vanhorn2018INaturalist, vanhorn2021Benchmarking}, and the TreeOfLife-10M dataset  \cite{stevens2024BioCLIP}, which extends iNat21 with images collected from the Encyclopedia of Life image bank \cite{EOL}, and insect images from the BIOSCAN-1M \cite{gharaee2024step} dataset.
However, citizen scientist collected data are often biased towards charismatic taxa, such as birds and plants, underrepresenting insects and other invertebrates.
To address this, some datasets, such as IP102 \cite{wu2019IP102}, INSECT \cite{badirli2021FineGrained}, Insect-1M \cite{nguyen2024InsectFoundation}, and Pest24 \cite{wang2020Pest24}  focus only on insects.
While uncontrolled field and citizen-scientist data might be useful for model pretraining as \cite{stevens2024BioCLIP} and \cite{jain2024Insect} show, the images do not represent realistic routine monitoring setups and the distribution shift to an operational setup might be significant. 

In contrast to uncontrolled images, camera traps and controlled in-situ imaging setups are already widely used in wildlife conservation and animal behavior analysis \cite{tuia2022Perspectives, oliver2023Camera, beery2021iWildCam, norouzzadeh2021deep}.
These methods are becoming more common for invertebrate monitoring. A recently released AMI dataset \cite{jain2024Insect} captures moths with an in-situ device. The overall dataset contains 2.5M images collected from the GBIF database, with 14,105 identified specimens from actual traps.
Camera traps have also been applied to monitor pollinators above flowering plants \cite{bjerge2024deep, bjerge2023Accurate, stark2023YOLO}, species stuck in sticky-paper \cite{keasar2024STARdbi, geissmann2022Sticky}, and organisms drifting in rivers \cite{deschaetzen2023Riverine}.
A challenge with in-situ images is that they often are multi-object images and rarely have fine-grained labels as identification at high taxonomic resolution from images alone is difficult and in many cases impossible for taxonomists \cite{arje2020Human}.

Imaging specimens in a lab setting allows separating collection from imaging, making it possible to image specimens in congruence with existing monitoring programs, where samples are collected for lab identification. Most laboratory datasets consider terrestrial insects, as they are easy to collect and highly diverse. 
Datasets, such as the ALUS \cite{schneider2022Bulk} and BIOSCAN \cite{gharaee2024step, gharaee2024BIOSCAN5M}, have been created by collecting large amounts of insects from Malaise traps. The latter dataset exhibits an impressive scale of 5M specimens from 324,411 classes, with accompanying DNA barcoding data.

In contrast to terrestrial species, aquatic macroinvertebrates are more challenging to collect in large quantities and have been studied less with computer vision methods \cite{milosevic2020Application}. \citet{simovic2024Automated} present a dataset of high-resolution aquatic invertebrate images, containing 16,650 images from 5,550 specimens and 90 classes, with multi-view images of each specimen.
The multi-view imaging system BIODISCOVER \cite{arje2020Automatic}, the same device we use in this study, has been also previously used to collect aquatic invertebrates datasets \cite{arje2020Human, hoye2022Accurate}, but they are significantly smaller than AquaMonitor and still suffer from selection bias and are evaluated closed-set.

A summary of the properties of the most relevant openly available existing datasets is provided in Table~\ref{tab:datasets}. We compare our dataset to large, self-imaged datasets that have potential for real-life monitoring purposes. Most of the self-imaged datasets are imaged in a lab setting, with the exception of the AMI dataset \cite{jain2024Insect} imaged with camera traps. The most notable datasets that contain images from public image repositories and are not limited to invertebrates are also included. The image count is larger than the number of specimens for sequence and multi-view datasets, and lower for datasets having multi-object images.

%% file: 3_dataset.tex
\section{AquaMonitor dataset}

\begin{figure*}[t]
    \centering
\includegraphics[width=0.9\linewidth]{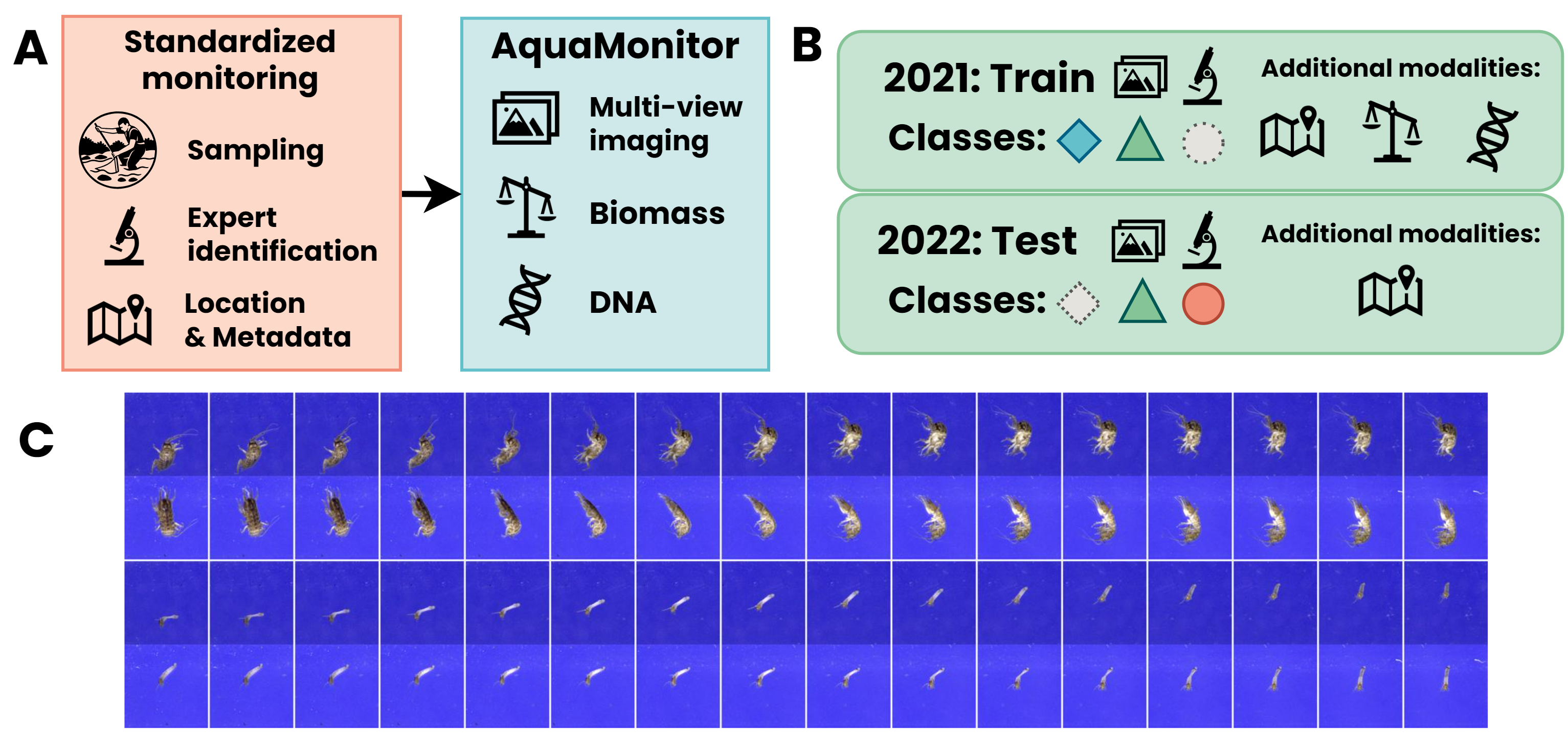}
    \caption{A: AquaMonitor was imaged in congruence with an operational routine monitoring program, ensuring high-quality sampling and identification. B: The monitoring benchmark uses 2021 data for training and 2022 for testing. C: Thumbnail examples of the synchronized multi-view sequences. Upper row species is the crustacean \textit{Asellus aquaticus}, lower row the caddisfly \textit{Mystacides azureus}.}
    \label{fig:overview}
\end{figure*}

The AquaMonitor dataset summarized in Fig.~\ref{fig:overview} consists of imaged samples from two years (2021 and 2022) of operational monitoring. 
The samples were collected from 50 sampling sites in 22 lakes.
The set of sites is different but partially overlapping for the two years due to the regular rotation of lakes in the monitoring program.
We imaged 44,854 multi-view image sequences from 43,189 benthic macroinvertebrate specimens, totaling to 2,756,664 images.
Examples of images can be seen in Fig.~\ref{fig:overview} C.
The 2021 data includes two mutually exclusive subsets containing DNA sequences and biomass information.
AquaMonitor has also rich metadata for each specimen, for example, its sampling location, and sampling and imaging times.
The numbers of images, specimens, image sequences, classes, lakes, and sampling sites in different years and subsets, as well the overlap over the years are summarized in Table~\ref{tab:overview}.
Lake names, numbers of sites, and specimen counts for each site can be found in the supplementary material Sec.~\ref{app:lakes}.

There are 152 different classes in a hierarchical taxonomic structure as illustrated in Fig.~\ref{fig:taxa}, including different life stages, such as juvenile and adult forms of some species. When considering only taxonomic groups, there are 145 taxa in total.  The number of taxa labeled to different hierarchical levels is given in Table~\ref{tab:taxalevels}.

\begin{table}
\centering
\begin{minipage}[]{.58\linewidth}
\centering
\small
\captionof{table}{\textbf{AquaMonitor statistics.} Overlap is the number of classes, lakes, and sampling sites common across both years.}
    \label{tab:overview}
    \resizebox{\linewidth}{!}{%
    \begin{tabular}{lrrrr}
    \toprule
    &2021	&2022	& Overlap	&Total\\
    \cmidrule{2-5}
    Images	& 1,640,936	&1,115,728	&	&2,756,664\\
    \hspace{3mm}{\small \textit{with DNA}}&{\small\textit{307,826}}&&&{\small\textit{307,826}}\\
    \hspace{3mm}{\small \textit{with Biomass}}&{\small\textit{120,627}}&&&{\small\textit{120,627}}\\
    Specimens	&22,882	&20,307	&	&43,189\\
    \hspace{3mm}{\small \textit{with DNA}}&{\small \textit{1358}}&&&{\small \textit{1358}}\\
    \hspace{3mm}{\small \textit{with Biomass}}&{\small \textit{1494}}&&&{\small \textit{1494}}\\
    Image sequences	& 24,547	&20,307&		&44,854\\
    \hspace{3mm}{\small \textit{with DNA}}&{\small \textit{2764}}&&&{\small \textit{2764}}\\
    \hspace{3mm}{\small \textit{with Biomass}}&{\small \textit{1582}}&&&{\small \textit{1582}}\\
    Classes	&128	&109	&85	&152\\
    \hspace{3mm}{\small \textit{with DNA}}&{\small \textit{23}}&&&{\small \textit{23}}\\
    \hspace{3mm}{\small \textit{with Biomass}}&{\small \textit{31}}&&&{\small \textit{31}}\\
    Lakes	&17	&13	&8	&22\\
    Sampling sites	&41	&29	&20	&50\\
    \bottomrule
    \end{tabular}}
\end{minipage}
\quad
\begin{minipage}[]{.38\linewidth}
    \centering
    \includegraphics[width=\linewidth]{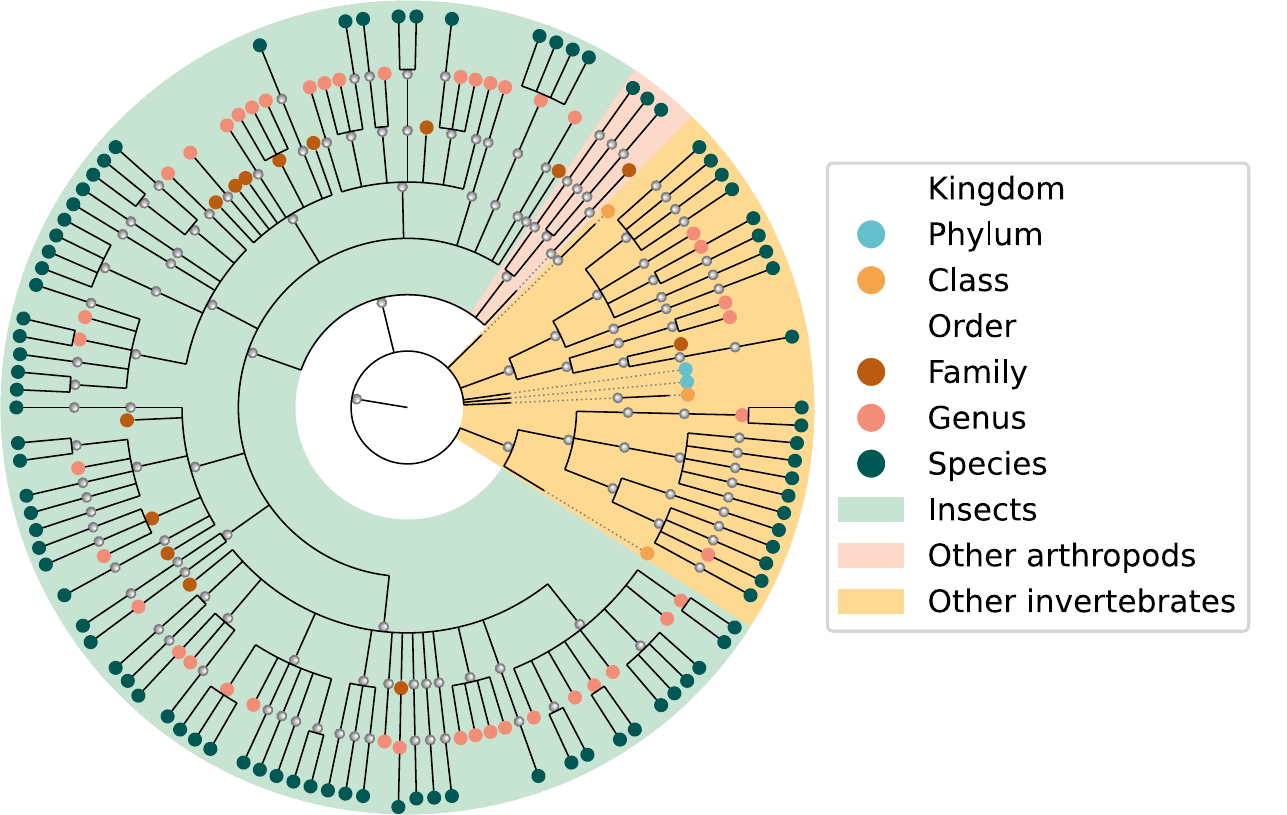}
    \vspace{3pt}
    \captionof{figure}{\textbf{Overview of the dataset specimen taxonomy.} The labels are hierarchical in nature, based on the GBIF backbone taxonomy. Colored nodes represent specimens labeled to this level. A detailed taxonomy with all scientific names are in the supplementary material Figures~\ref{fig:ept},~\ref{fig:non-ept} and~\ref{fig:noninsects}}
    \label{fig:taxa}
\end{minipage}

\end{table}

\begin{table}
\small
\captionsetup{font=footnotesize}
\centering
\begin{minipage}[]{.48\linewidth}
    \centering
    \scriptsize
    \captionof{table}{\textbf{The number of taxa labeled to different hierarchical levels.} Unique column shows the number of taxonomic groups, when the most accurate taxonomic rank is set to the this level. Variations refer to different life stages, such adult and juvenile forms, of some taxa.}
    \label{tab:taxalevels}
    \resizebox{\linewidth}{!}{%
    \begin{tabular}{lcp{1.2cm}c}
    \toprule
    Group &Unique	& Labeled to \newline this level	&Variations	\\
    \midrule
    Kingdom&	1&	0&\\	
    Phylum&	7&	2&\\	
    Class&	11&	3&\\	
    Order&	25&	0&\\	
    Family&	63&	14&\\	
    Genus&	110&	37&	6\\
    Species&	145&	89&	1\\
    \bottomrule
    \end{tabular}}
\end{minipage}
\quad
\begin{minipage}[]{.40\linewidth}
    \centering
    \captionof{table}{\textbf{Benchmark split statistics.} The classification and few-shot tasks use five cross-validation folds, these values being from the first fold.}
    \label{tab:benchmark_overview}
    \resizebox{\linewidth}{!}{%
    \begin{threeparttable}
    \begin{tabular}{lrrr}
    \toprule
                        & Train & Val & Test \\
                    \midrule
                    \multicolumn{4}{c}{Monitor} \\
                    \midrule
                    Images & 1,640,936                   & 110,207                  & 1,005,521                  \\
                    Specimens & 22,882                     & 2028                    & 18,279                    \\
                    Classes   & 128                         & 60                      & 85+24$^\star$                    \\
                    \midrule
                    \multicolumn{4}{c}{Classification} \\
                    \midrule
                    Images & 1,882,046                   & 282,923                  & 543,054                   \\
                    Specimens & 29,346                     & 4386                    & 8433                     \\
                    Classes   & 42                         & 42                      & 42                    \\
                    \midrule
                    \multicolumn{4}{c}{Few-shot} \\
                    \midrule
                      Images & 30,575                     & 4426                    & 8382                     \\
                    Specimens & 487                       & 60                      & 145                      \\
                    Classes   & 47                         & 32                      & 47                    \\
                    \bottomrule
    \end{tabular}
    \begin{tablenotes}
    \footnotesize
    \item $\star$ 85 in-distribution + 24 out-of-distribution classes
    \end{tablenotes}
    \end{threeparttable}}
\end{minipage}\hfill
\end{table}

\subsection{Benchmark tasks}
\label{sec:benchmarks}
We define three different benchmark tasks on the dataset: \textbf{monitoring}, \textbf{classification}, and \textbf{few-shot}.
The \textbf{monitoring} benchmark is the most important, as it includes all the challenges encountered in an operational monitoring setting, and performance on this benchmark can directly translate to routine monitoring efforts.
The benchmark uses all specimens from 2021 for training and specimens from 2022 for validation and testing, which also reflects a realistic scenario that could be followed in future monitoring efforts.
The validation set is randomly selected 10\% of the 2022 specimens.
The class distribution is extremely imbalanced, with 63 classes having under 5 specimens, and 22 classes having only a single example.
The test set has 85 in-distribution (ID) classes common with the training set and 24 out-of-distribution (OOD) classes. The goal of the benchmark is to classify the ID classes as accurately as possible and to detect the OOD classes reliably.

The \textbf{classification} benchmark groups both years together and uses a subset of 42 classes that have at least 50 specimens. This setup follows the standard fine-grained classification task in a closed-set setting and without domain shift. The benchmark aims at showing the difference between the monitoring benchmark and this kind of setup typically used in the currently available datasets. The data used for the \textbf{few-shot} benchmark consists of 47 classes with only 5-49 specimens, falling in the few-shot learning domain, where standard classification approaches might not work. A good model should learn robust representations and generalize to new classes using only a few examples.

We provide predefined train-test-val splits across five cross-validation folds for the classification and few-shot tasks. Splits are stratified by taxa, sampling site, and the presence of DNA and biomass metadata, making it possible to use the same splits with different subsets. Benchmark split statistics are shown in Table~\ref{tab:benchmark_overview}. More details on the splits are given in the supplementary material Sec.~\ref{app:splits}.

\subsection{Dataset collection}

\subsubsection{Sampling protocol and taxonomic identification}

The specimens in the AquaMonitor dataset are from a nationwide freshwater monitoring program, that has monitored the effects of agriculture and forestry on water bodies in Finland since 2008 \cite{vilmi2021maa}.
The specimens were collected from lake shores, following the EU \ac{WFD} kick-sampling protocol, and were stored in 96\% ethanol after sampling.
Each lake has 1-3 sites, which were sampled during September and October in 2021 and 2022. Details on sampling are in the supplementary material Sec.~\ref{app:sampling}.

The samples were morphologically identified by expert taxonomists as part of the monitoring program.
The identification was carried out individually for each specimen using a microscope and standard tools.
Specimens were classified to the lowest feasible level.
Most of the taxa (89/145) were identified to species level, but many specimens were identified to higher levels of taxonomic hierarchy, since identification to lower levels is inherently difficult.
For example, nematodes (eelworms) were classified only down to the phylum level.
A challenging property of the taxonomy is that not all classes are leaf nodes of the taxonomy.
For example, the dataset has caddisfly specimens on three levels: on family (Limnephilidae), on genus (\textit{Limnephilus}), and on species (\textit{Limnephilus pantodapus}) level.
Taxon groups with child nodes follow a convention, where the higher level group contains only specimens that were not possible to identify on lower levels.  
Thus, classes are mutually exclusive - an important property for building classifiers.

The sampling and species identification were carried out by the monitoring program, independently from this study.
Freshwater monitoring programs in Finland follow strict protocols, ensuring high-quality sample collection and species identification.
We received the samples after expert identification, sorted into containers by taxon and sampling site.
Metadata on sampling locations and times were obtained from a national monitoring database, where species observation records are collected.
Fig.~\ref{fig:overview}A illustrates the division of responsibilities between the monitoring program and our study.

\subsubsection{Imaging and imaging coverage}

We used the BIODISCOVER device \cite{arje2020Automatic} for imaging. During imaging, each specimen was dropped into an 1cm $\times$ 1cm $\times$ 3.5cm cuvette filled with 91\% ethanol.
As the specimen falls through the cuvette, a sequence of images is captured from two perpendicular Basler acA1920-155uc cameras.
We used an aperture of f/8, an exposure time of 2000 microseconds, and a frame rate of 50 frames per second.
The DNA subset was imaged with 96\% ethanol leading to different falling speeds for this subset.

We imaged all the specimens that we received and were feasible to image using our imaging setup.
This does not mean that every specimen collected in the official monitoring program was imaged, as some samples had to be used for other purposes, were lost during transportation and handling or were not suitable for imaging using the BIODISCOVER device.
Comparing our specimen counts to the monitoring database, we were able to image 89.58\% (out of 25,546) of 2021 specimens and 72.65\% (out of 27,952) of 2022 specimens.
A large part of missing 2022 specimens were from 6 lakes we were not able to get any specimens from. The taxonomic coverage of our dataset is 152 taxa out of 161 taxa encountered during the two monitoring years. The missing taxa were either too big to fit in the imaging device or too small for the camera to detect.
A list of the missing taxa and additional details on dataset coverage are provided in the supplementary material Sec.~\ref{app:coverage}.

The imaging setup captures multi-view image sequences for each specimen.
All specimens have sequences from two views, except for 222 which have only one view due to camera malfunctions, resulting in 89,474 sequences from different views. If the specimen is smaller than the width of the cuvette, the saved image is square, which is the case for 99\% of images. Most images are of resolution 464x464px, and at least 412px on the shortest side. The longer side can be up to 1114px for large specimens.
The position of the image crop was saved. This makes it possible to calculate metrics, such as falling speed, for each specimen.
The average number of images per specimen is 63 (IQR 40-73). The sequence length correlates with the weight of the specimen - the heavier the specimen is, the faster it falls, and less images are captured. The specimens in the DNA subset were imaged at least twice, with the goal of having at least 50 images per specimens for this subset. Some of the biomass specimens were also imaged twice. This results in slightly more imaging sequences than specimens as shown in Table~\ref{tab:overview}.

\subsubsection{Biomass and DNA subsets}
\textbf{Biomass.} The biomass subset specimens were measured and imaged using a digital microscope. We measured specimen length and head width, two measurements commonly used in biomass estimation \cite{wardhaugh2013Estimation}. We saved high-resolution images captured during this process (example images in the supplementary material Fig.~\ref{fig:hi-res}). After measurement, the specimens were dried in a drying oven for 22-24 hours in 105 degrees Celcius. After drying, the specimens were placed in a vacuum exicator before weighing to prevent moisture collection. The weighing was done using a precision scale with a precision of 0.5 $\mu g$. More details of the biomass subset can be found in the supplementary material Sec.~\ref{app:biomass}.

\textbf{DNA sequencing.} DNA barcoding is a frequently used method for species identification based on DNA sequences of a so called \textit{marker gene}, such as the mitochondrial cytochrome c oxidase subunit I (COI) gene. After extracting and sequencing the DNA of a specimen, it can be compared to a reference database \cite{ratnasingham2007BOLD} to provide an identification often to species level. %
Using fwhF2/FwhR2n primers \cite{vamos2017Short}, we sequenced a 205 bp long fragment of the COI marker gene of 1518 specimens from 23 classes and obtained DNA sequences for 1358 specimens. Details on laboratory work and this subset can be found in the supplementary material Sec.~\ref{app:dna}.

\subsection{What makes AquaMonitor dataset unique?}
\label{sec:what_makes_dataset_unique}

\textbf{Real-life monitoring setup}: AquaMonitor is the first aquatic invertebrate dataset that has been collected in congruence with an operational monitoring program. The dataset represents the full diversity of species encountered during regular biomonitoring, avoiding selection bias.
Although some datasets of terrestrial macroinvertebrates, including the self-imaged part of AMI \cite{jain2024Insect} dataset, ALUS Southern Ontario dataset \cite{schneider2022Bulk}, and BIOSCAN-5M \cite{gharaee2024BIOSCAN5M}, have been collected in an operational manner, they do not mention or give statistics of any monitoring programs.

\noindent\textbf{Multi-view image sequences}: A feature of our dataset is synchronized multi-view image sequences of each specimen, where the specimen is imaged simultaneously using two perpendicular cameras. Only other species datasets with this property were also collected using the BIODISCOVER imaging device \cite{arje2020Automatic, arje2020Human, hoye2022Accurate}. The largest previous dataset consists of 9631 individuals from 39 classes, totaling 460,009 frames, being significantly smaller than our dataset.
Multi-view sequences make it possible to use AquaMonitor for generic fine-grained multi-view object classification tasks.
There is a clear lack of benchmark datasets for this task, with only few datasets available in general \cite{su2015MultiView, hamdi2021MVTN, seeland2021Multiview, mader2021Flora, marques2018Ant, held2023VARS, vyas2020Multiview, vu2024MmCows}.

\noindent\textbf{Rich metadata}: AquaMonitor includes sampling locations and times for each specimen. It also has DNA, biomass, and size information for subsets of images. 
Few biodiversity datasets contain any metadata in addition to the images. BIOSCAN-5M \cite{gharaee2024BIOSCAN5M} is the currently the only image-DNA dataset with self-collected and sequenced DNA. 
Although there are smaller biomass image datasets \cite{arje2020Automatic, wuhrl2022DiversityScanner, schneider2022Bulk}, AquaMonitor contains the largest number of individually measured biomass and size data for invertebrates.
While the evaluation in this paper focuses mainly on image-based benchmarks, DNA and biomass information creates opportunities for future research, for example, by further developing methods such as CLIBD \cite{gong2024CLIBD} or DNA-based OOD detection \cite{impio2024Improving}.

\noindent\textbf{Label granularity}: 
Morphological identification is challenging for many species, and often requires a microscope and inspection of the physical specimen. Accordingly, AquaMonitor samples were identified by a professional taxonomist using a microscope, with strict quality assurance. Many previous datasets have struggled to label specimens consistently. BIOSCAN \cite{gharaee2024BIOSCAN5M} and AMI dataset \cite{jain2024Insect} species are identified from images, thus reducing the feasible depth of identification.

%% file: 4_experiments.tex
\section{Benchmark experiments and results}
\label{sec:results}

\subsection{Experimental setup} 
\textbf{Monitoring benchmark:} We trained two variants from four common backbone classes: ResNets (50, 101) \cite{he2016Deep}, EfficientNets (B0, B4) \cite{tan2019EfficientNet}, Vision transformers (ViT-B/16, ViT-L/14) \cite{dosovitskiy2021Image}, and Swin transformers (Swin-T, Swin-B) \cite{liu2021Swin}, as well as a single MobileNetV3 model \cite{howard2019Searchinga} for reference.
We also trained two models derived from BioCLIP, which is a generic species classification model trained with the TreeOfLife-10M dataset \cite{stevens2024BioCLIP}: one with full fine-tuning and another one with only the last two transformer blocks and the classification head being trainable. Based on initial evaluation on the validation set, we chose Swin-T as the backbone for a \textbf{multi-view} model that uses image inputs from both cameras. 
We performed evaluation also on an \textbf{ensemble model} that combines the EfficientNet-B4, Swin-T, and multi-view models. Details on the multi-view model architecture and the ensemble can be found in the supplementary material Sec.~\ref{app:experimental_setup}.

All other models except EfficientNet-B4 were trained for 100 epochs with the AdamW optimizer \cite{loshchilov2019Decoupled}, using an initial learning rate of 0.0001 and a cosine annealing learning rate scheduler \cite{loshchilov2017SGDR}. The EfficientNet-B4 suffered from severe overfitting and was trained for only 20 epochs.
For data augmentation, we used TrivialAugment \cite{muller2021TrivialAugment}.
All inputs were resized to 224x224, except for EfficientNet-B4, which uses 320x320 inputs.
Models were trained using the Lightning framework \cite{falcon2019PyTorch}, using pretrained weights \cite{russakovsky2015ImageNet, radford2021learning} from the \texttt{timm}-library \cite{rw2019timm}.
Details on pretrained weights, specific models, and computational resources used are given in the supplementary material Sec.~\ref{app:experimental_setup}.

The monitoring benchmark has two tasks: \textbf{in-distribution classification} and \textbf{out-of-distribution detection}.
For classification, image sequence information was used by classifying each sequence frame separately and averaging all logit outputs for a specimen. The maximum logit class was chosen as the final prediction. For OOD, we used ranking-based approaches, which are strong baselines. We used ranking metrics of entropy, MaxLogit \cite{hendrycks2022Scaling}, and Energy \cite{liu2020Energybased}.

\textbf{Classification benchmark:} Following results from the monitoring benchmark, we trained new models on the standard classification task with a closed set of classes, enough training examples and no domain shift to show how much easier it is compared to the monitoring benchmark.
The architectures chosen for the classification task were EfficientNet-B0 and B4, ResNet50, Swin-T, and MobileNetV3, as well as a similar Swin-T multi-view model and an ensemble classifier as in the Monitoring benchmark.
The models were trained with the same training protocols as above.

\textbf{Few-shot classification benchmark:} We used a simple 5-nearest-neighbors baseline for the few-shot classification task. The nearest neighbor search was done by cosine distance between image embeddings. We used the classification models above as the feature encoders. We also tested pretrained CLIP \cite{radford2021learning}, DINO \cite{caron2021Emerging, oquab2023DINOv2}, SigLIP \cite{zhai2023Sigmoid, tschannen2025SigLIP} and BioCLIP \cite{stevens2024BioCLIP} models as a training-free approach.
Since the few-shot dataset is significantly smaller than the classification dataset, we evaluated the results across all five cross-validation folds. The test sets for these folds are mutually exclusive and are pooled together after prediction in a jackknife-manner.

\textbf{Biomass estimation:} We trained regression models for 50 epochs using the Swin-T backbone and the same optimizer setup as above. We used four training variations to illustrate the importance of domain-specific representations for this task: two models started from ImageNet weights and two from the classification models above. We trained models both having feature encoders frozen and having them trainable. As we observed regression task performance to be very sensitive to the applied learning rate, we ran a short learning rate search for each model. The objective function was the mean absolute error between log-transformed biomass values and the model outputs.

\begin{table}[htb]
\centering
\begin{minipage}[]{.48\linewidth}
    \centering
    \caption{\textbf{Monitoring benchmark results.} The monitoring dataset was evaluated with 85 in-distribution classes. Full table with computational requirements and bootstrapped 2-sigma error bars are in the supplementary material Table~\ref{tab:classif-fulltable}.}
    \label{tab:monitor_results}
    \resizebox{\linewidth}{!}{%
    \begin{tabular}{lrrrr}
    \toprule
    Model & Accuracy & Top-5 & F1 macro & F1 weighted \\
    \midrule
    MobileNetV3 & 0.751 & 0.941 & 0.228 & 0.713 \\
    ResNet-50 & 0.851 & 0.963 & 0.327 & 0.826 \\
    ResNet-101 & 0.857 & 0.959 & 0.322 & 0.834 \\
    EfficientNet-B0 & 0.856 & 0.972 & 0.305 & 0.836 \\
    EfficientNet-B4 & 0.867 & 0.965 & 0.315 & 0.843 \\
    Swin-T & 0.870 & 0.985 & 0.361 & 0.850 \\
    Swin-T (Multiview) & 0.879 & 0.981 & 0.338 & 0.858 \\
    Swin-B & 0.858 & 0.980 & 0.335 & 0.838 \\
    ViT-B/16 & 0.811 & 0.969 & 0.292 & 0.800 \\
    ViT-B/16 (BioCLIP) & 0.817 & 0.960 & 0.258 & 0.790 \\
    ViT-B/16 (BioCLIP-FT) & 0.835 & 0.968 & 0.326 & 0.809 \\
    ViT-L/14 & 0.854 & 0.975 & 0.315 & 0.837 \\
    Ensemble & 0.882 & 0.984 & 0.367 & 0.859 \\
    \bottomrule
    \end{tabular}}
\end{minipage}
\quad
\begin{minipage}[]{.48\linewidth}
    \centering
    \caption{\textbf{Selected classification and few-shot results.}}
    \label{tab:classif_results}
    \resizebox{\linewidth}{!}{%
    \begin{tabular}{lrrrrr}
    \toprule
    Model & Accuracy & Top-5 & F1 macro & F1 weighted \\
    \midrule
    \multicolumn{5}{c}{Classification results} \\
    \midrule
    MobileNetV3 & 0.969 & 0.997 & 0.904 & 0.968 \\
    ResNet-50 & 0.981 & 0.998 & 0.926 & 0.980 \\
    EfficientNet-B0 & 0.983 & 0.998 & 0.933 & 0.983 \\
    EfficientNet-B4 & 0.985 & 0.999 & 0.944 & 0.985 \\
    Swin-T & 0.988 & 0.999 & 0.945 & 0.988 \\
    Swin-T (Multiview) & 0.985 & 0.998 & 0.937 & 0.985 \\
    Ensemble & 0.988 & 0.999 & 0.948 & 0.988 \\
    \midrule
    \multicolumn{5}{c}{Few-shot results} \\
    \midrule
    EfficientNet-B4 & 0.828 & 0.942 & 0.779 & 0.821 \\
    Swin-T & 0.829 & 0.936 & 0.781 & 0.822 \\
    \midrule
    CLIP/BioCLIP & 0.759 & 0.937 & 0.720 & 0.754 \\
    CLIP/OpenAI & 0.723 & 0.918 & 0.650 & 0.711 \\
    DINO & 0.758 & 0.913 & 0.709 & 0.753 \\
    SigLIP & 0.733 & 0.937 & 0.655 & 0.726 \\
    \bottomrule
    \end{tabular}}
\end{minipage}
\end{table}

\subsection{Experimental results}

\noindent\textbf{Monitoring task} in-distribution classification results are given in Table~\ref{tab:monitor_results} and Fig.~\ref{fig:monitor_classwise}, with a detailed confusion matrix in the supplementary material Fig.~\ref{fig:confusion-monitor}.
The best performing models were the multi-view and single-view Swin-T models, and EfficientNet-B4.
Combining these to an ensemble model produces the overall best model.
We observe that Swin models perform better than ViT models.
Using BioCLIP weights gives only slightly better results than a plain ViT model with regular CLIP weights. We observe that even though overall accuracy is high, performance on many classes remains low.

\begin{figure*}[htb]
\centering
\begin{minipage}[]{.4\linewidth}
    \centering
    \includegraphics[width=\linewidth]{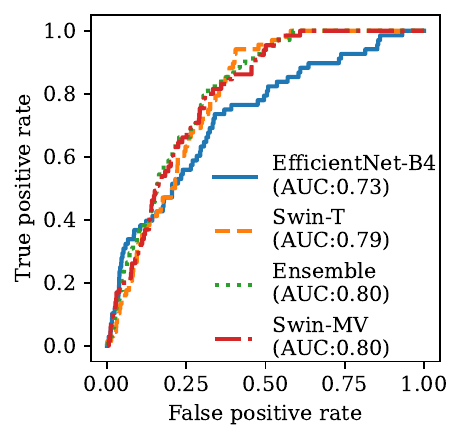}
    \captionof{figure}{\textbf{OOD detection ROC curve} Out-of-distribution detection on the 72 specimens belonging to 24 OOD classes, using the MaxLogit OOD scoring metric \cite{hendrycks2022Scaling}.}
    \label{fig:ood}
\end{minipage}
\quad
\begin{minipage}[]{.48\linewidth}
    \centering
    \captionof{table}{\textbf{Biomass estimation} with different pretraining datasets, evaluated using mean absolute error (MAE), and median and mean absolute percentual errors (MdAPE, MAPE). For frozen models, we trained only the final projection layer.}
    \label{tab:biomass}
    \resizebox{\linewidth}{!}{%
    \begin{tabular}{cc|rrr}
    \toprule
    Dataset & Frozen & MdAPE & MAE & MAPE \\
    \midrule
    ImageNet & \checkmark & 0.686 & 0.251 & 1.858 \\
    ImageNet &  & 0.196 & 0.120 & 0.457 \\
    AquaMonitor & \checkmark & 0.241 & 0.144 & 0.582 \\
    AquaMonitor & & 0.173 & 0.113 & 0.431 \\
    \bottomrule
    \end{tabular}}
    
    \captionof{table}{\textbf{Multi-view sequence effects.} The amount of available data is seen in model performance. 1C: single camera, 2C: both cameras. MV: multi-view.}
    \label{tab:multiview}
    \resizebox{\linewidth}{!}{%
    \begin{threeparttable}
    \begin{tabular}{lccc}
        \toprule
        Method     & F1 Macro & F1 Weighted & Accuracy \\
        \midrule
        Image      & 0.281    & 0.808       & 0.826    \\
        Sequence, 1C   & 0.263    & 0.823       & 0.841    \\
        Sequence, 2C$^\star$ & 0.361    & 0.850       & 0.870    \\
        MV model$^\dagger$ & 0.338    & 0.858 &  0.879 \\
        \bottomrule
    \end{tabular}
    \begin{tablenotes}
    \footnotesize
    \item $\star$ equals Swin-T in Table~\ref{tab:monitor_results}
    \item $\dagger$ equals Swin-T multi-view in Table~\ref{tab:monitor_results}
    \end{tablenotes}
    \end{threeparttable}}
\end{minipage}
\end{figure*}

OOD detection performance, shown as a ROC curve in Fig.~\ref{fig:ood} for best performing MaxLogit ranking metric, shows that OOD detection is challenging using common ranking-based approaches.
OOD detection results for entropy and energy scores are provided in the supplementary material Table~\ref{tab:ood-appendix}.

\textbf{Classification benchmark} results are given in Table~\ref{tab:classif_results}, with more detailed confusion matrices and class-wise results in the supplementary material Figures~\ref{fig:confusion} and~\ref{fig:clf_fewshot_classwise}.
We observe that performance in standard classification is good, which is in line with previous observations showing that closed-set fine-grained classification tasks are fairly easy when enough data are available \cite{schneider2023Getting}. Even the large imbalance in the data does not hurt the performance significantly. Full results for all trained models are in the supplementary material Table~\ref{tab:classif-fulltable}.

\textbf{Few-shot classification} results are also given in Table~\ref{tab:classif_results}.
K-NN classification performance using feature encoders trained with AquaMonitor data perform better, as expected.
However, using pretrained models works moderately well.
The BioCLIP model, trained with species data, works the best, but DINO features are very close although not being trained explicitly on species data. Full results with worse performing DINOv2 and SigLIP2 models are in the supplementary material Table~\ref{tab:fewshot-fulltable}.

\textbf{Biomass estimation} results are given in Table~\ref{tab:biomass}. We can observe that features learned from the full classification task carry out to the biomass estimation task significantly better than ImageNet features. Regression scatter plots for biomass estimation can be found in the supplementary material Fig.~\ref{fig:biomass-scatter}.

\textbf{Multi-view and sequence analysis} is given in Table~\ref{tab:multiview}. The table illustrates Swin-T performance on the monitoring benchmark, with increasing amounts of data. Adding sequence information improves overall performance slightly, but using sequences from both cameras yields a larger gain. The multi-view model produces again a slight improvement.

\begin{figure*}[htb]
    \centering
    \includegraphics[width=\linewidth]{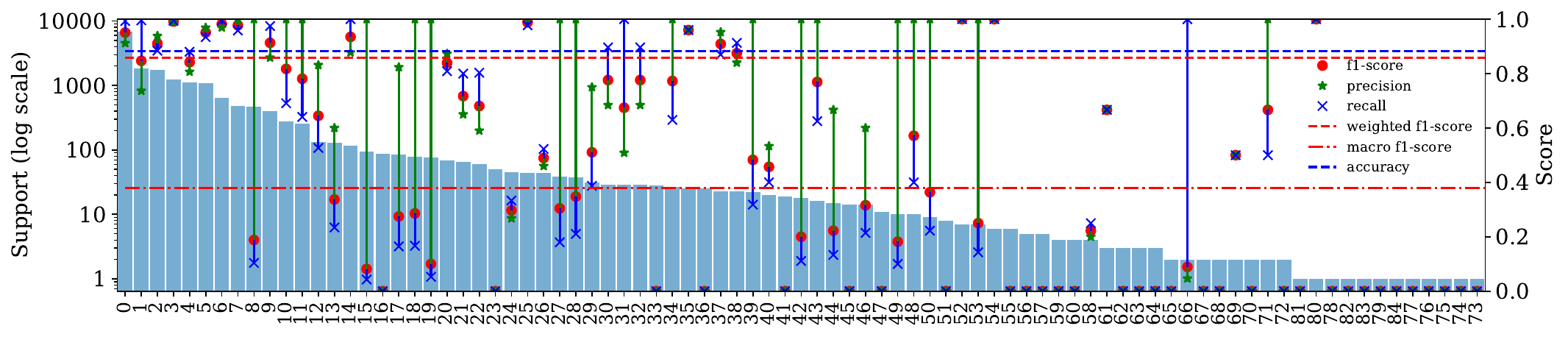}
    \caption{\textbf{Class-wise accuracy of the ensemble model in the monitoring task.} Although overall performance is high (weighted F1: 0.859), many classes remain challenging. Taxon names referenced by numbers and a full result table can be found in the supplementary material Table~\ref{tab:monitor-fulltaxa}.}
    \label{fig:monitor_classwise}
\end{figure*}

%% file: 5_conclusions.tex
\vspace{-3mm}
\section{Conclusions}
\label{sec:conclusions}

We presented the AquaMonitor dataset, a new multi-view image sequence dataset of aquatic invertebrates. The dataset is a valuable resource for evaluating computer vision methods for biodiversity monitoring tasks, as well as computer vision tasks, such as ultra-fine-grained visual categorization, imbalanced classification, few-shot learning, and open-set recognition.

The main limitations of our study are that the dataset contains specimens from only a single country, and that evaluations lack multimodal fusion approaches. This study was deemed not to have room for complex multimodal approaches. The dataset successfully covers the diversity of Finnish lake invertebrates, which is in fact quite low compared to many countries. Finnish species are also relatively well-known, which is not the case in many countries with a lot of undescribed species. International collaboration and integration of molecular methods will be needed to capture species diversity around the world and improve these methods further.

\textbf{Acknowledgements}. AquaMonitor dataset was compiled under Research Council of Finland project 333497. The paper and some of the experiments were finalized with the support of Finnish Research Infrastructure (FIRI) funding instrument FinBIF FIRI 345733 and Biodiversa+ BiodivMon project DNAquaIMG VN/29767/2023-YM-7. We thank Jukka Aroviita and the MaaMet project for providing the specimens, Terhi Lensu for the exceptional work in species identification, Gabriel Reichert, Riku Karjalainen, and Pirjo Appelgren for the imaging work, and CSC – IT Center for Science, Finland, for computational resources.

%% file: X_suppl.tex
\clearpage
\setcounter{page}{1}

\section{Supplementary material for Section 3  AquaMonitor dataset}

\subsection{Lakes and sites}
\label{app:lakes}

\begin{figure}[htb]
    \centering
\includegraphics[width=0.5\linewidth]{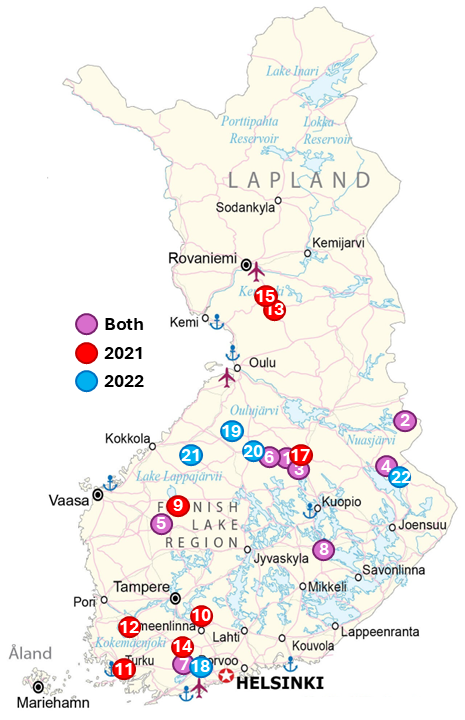}
    \caption{Sampling lake locations.}
    \label{fig:map}
\end{figure}

Species counts per site are given in Tables~\ref{tab:sites_appendix} and~\ref{tab:sites_appendix2}. Table~\ref{tab:sites_appendix} shows lakes that are common for both sampling years. Table~\ref{tab:sites_appendix2} shows lakes that have specimens from only one year.
The lakes do not fully overlap between years due to rotation in the lakes. This is a part of the monitoring program.
The approximate locations of the lakes on the map are shown in Fig.~\ref{fig:map}, with numbers corresponding to the numbers in the tables.

\subsection{Label taxonomy}
\label{app:taxa}

\begin{figure*}[htb]
    \centering
\includegraphics[width=0.55\linewidth]{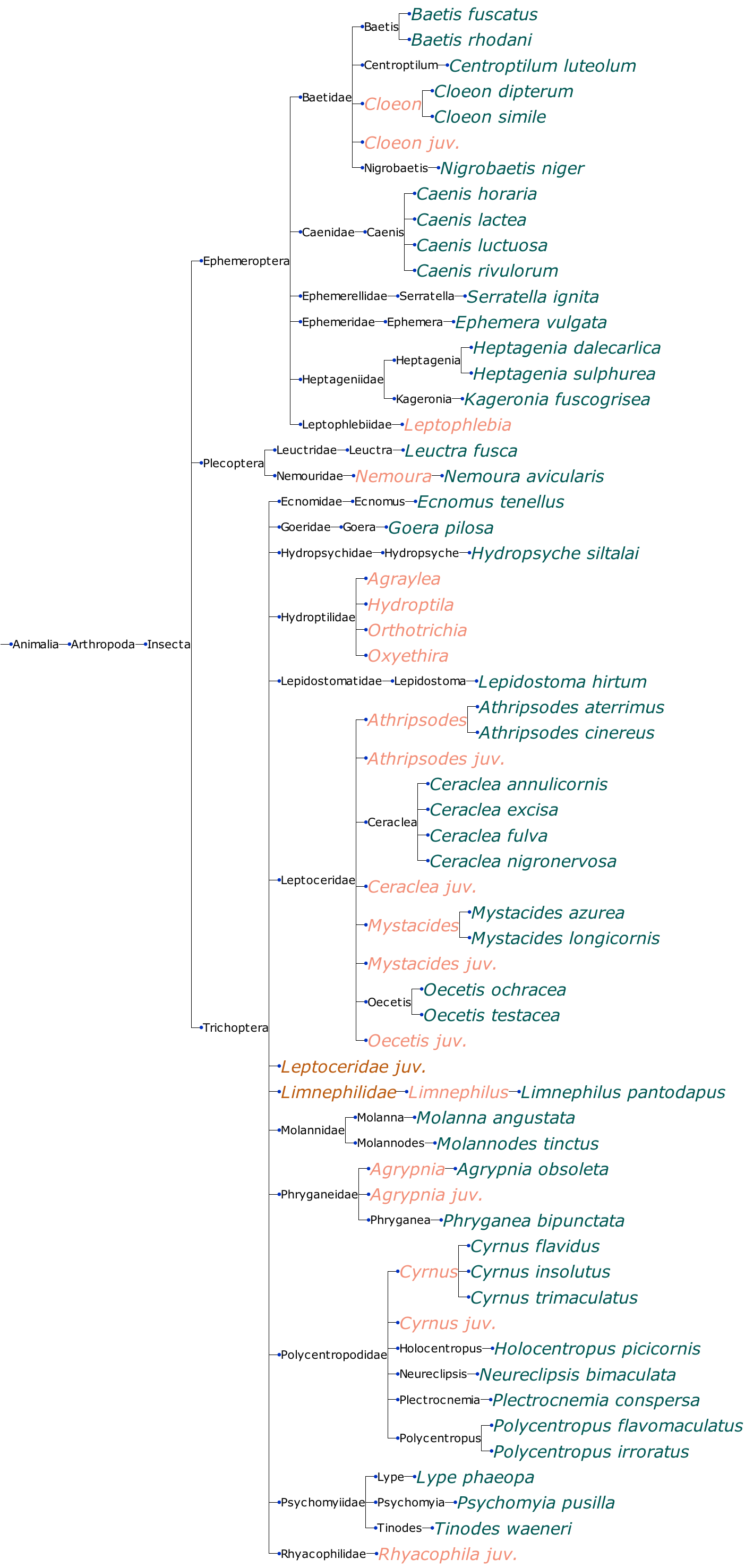}
    \caption{Insects from orders Ephemeroptera, Plecoptera, and Trichoptera.}
    \label{fig:ept}
\end{figure*}

\begin{figure*}[htb]
    \centering
\includegraphics[width=0.6\linewidth]{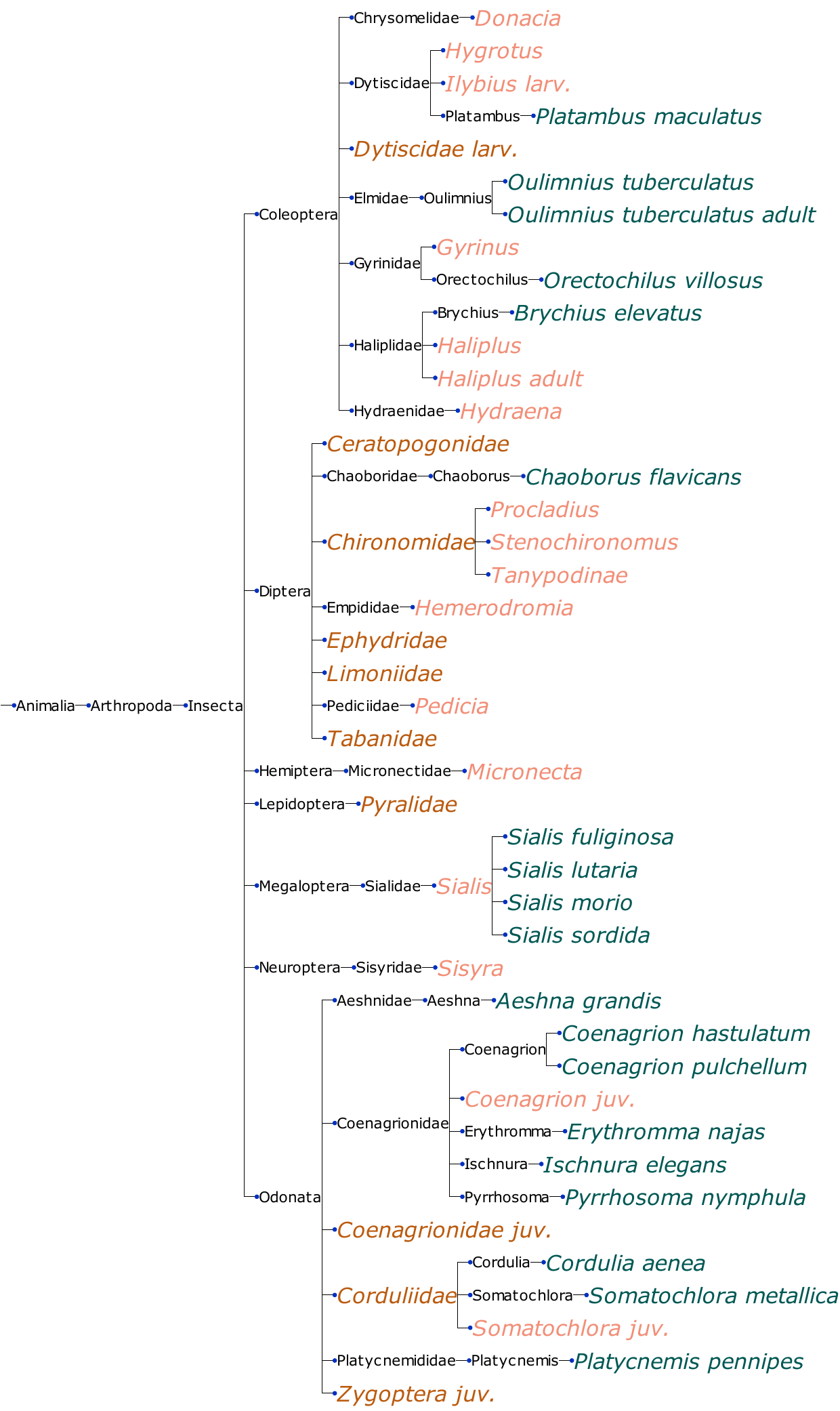}
    \caption{Insects other than EPT taxa.}
    \label{fig:non-ept}
\end{figure*}

\begin{figure*}[htb]
    \centering
\includegraphics[width=0.6\linewidth]{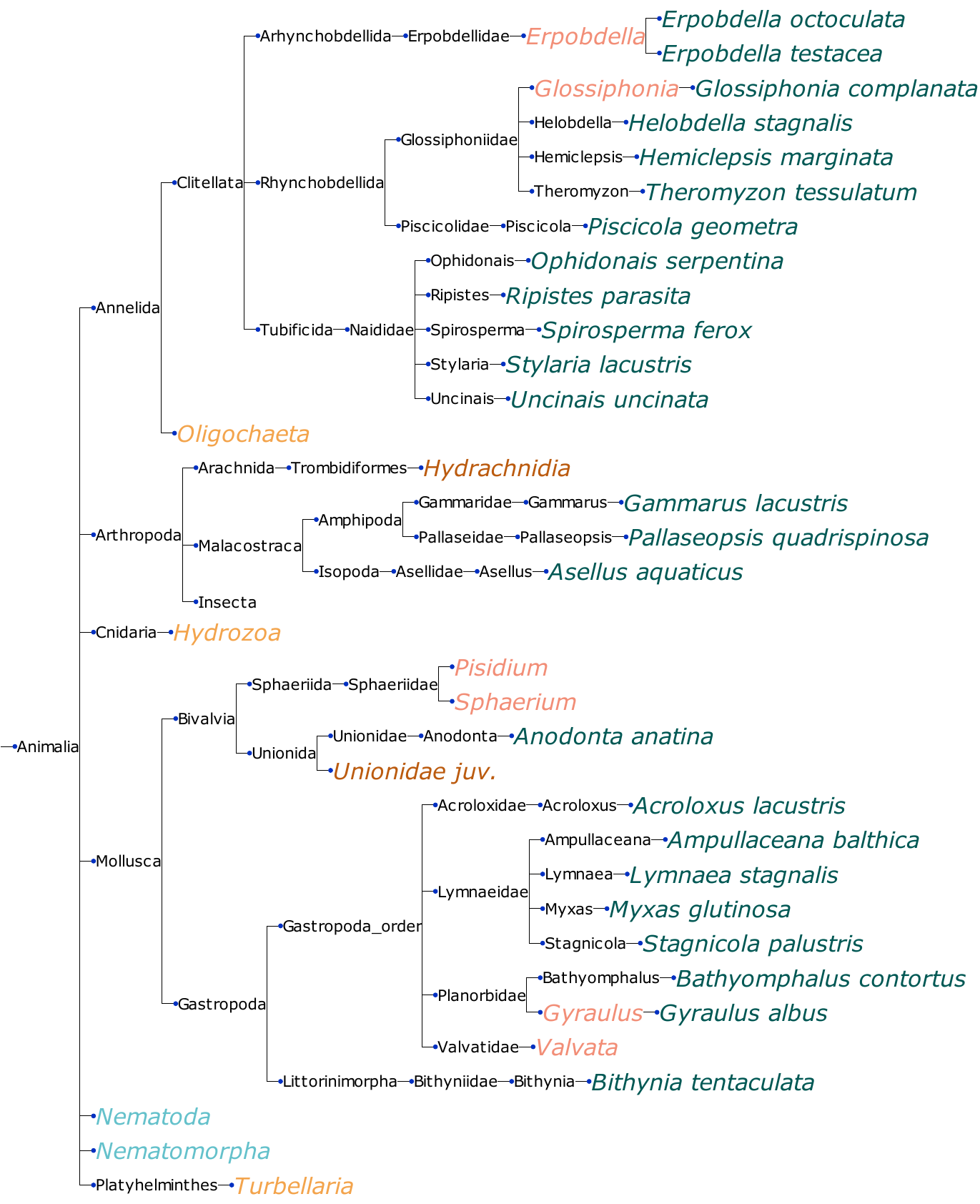}
    \caption{Non-insects.}
    \label{fig:noninsects}
\end{figure*}

Fig.~\ref{fig:ept}, Fig.~\ref{fig:non-ept}, and Fig.~\ref{fig:noninsects} show Fig.~2 from the main paper in full detail. The taxonomic tree is divided into three parts for illustrative purposes: EPT taxa (insects from taxonomic orders Ephemeroptera, Plecoptera, and Trichoptera, which are important groups in aquatic monitoring), other insects, and all other invertebrates. The colors correspond to the colors used in Fig.~2 in the main paper. Colored and cursive text indicates that specimens with this label are present in the dataset. Fig.~\ref{fig:image-example-all} shows one randomly chosen example image of each of the 152 classes. Fig.~\ref{fig:heatmap} illustrates the image counts of each taxonomic family and sampling site pair. The image counts of all descendant classes are summed together for each family. Some taxa are present on almost all sites, but some taxa are more rare.

\begin{table}[htb]
\caption{Specimen counts for lakes with samples from both years.}
\centering
\small
\label{tab:sites_appendix}
\begin{tabular}{llrr}
\toprule
Lake & Site & 2021 & 2022 \\
\midrule
\multirow[t]{3}{*}{1 Haapajärvi} & haa1 & 163 & 193 \\
 & haa2 & 119 & 422 \\
 & haa3 & 168 & 397 \\
\midrule
\multirow[t]{3}{*}{2 Iso Riihijärvi} & iso1 & 2457 & 1393 \\
 & iso2 & 1712 & 1180 \\
 & iso3 & 3942 & 996 \\
\midrule
\multirow[t]{2}{*}{3 Kirmanjärvi} & kir1 & 169 & 185 \\
 & kir2 & 113 & 114 \\
\midrule
\multirow[t]{2}{*}{4 Kuohattijärvi} & kuo1 & 215 & 587 \\
 & kuo2 & 582 & 503 \\
\midrule
\multirow[t]{2}{*}{5 Kuortaneenjärvi} & kur1 & 409 & 452 \\
 & kur2 & 431 & 468 \\
\midrule
\multirow[t]{2}{*}{6 Niemisjärvi} & nie1 & 142 & 279 \\
 & nie2 & 373 & 440 \\
\midrule
\multirow[t]{3}{*}{7 Pusulanjärvi} & pus1 & 467 & 100 \\
 & pus2 & 317 & 148 \\
 & pus3 & 538 & 458 \\
\midrule
\multirow[t]{3}{*}{8 Valvatus} & val1 & 349 & 1319 \\
 & val2 & 335 & 1173 \\
 & val3 & 215 & 510 \\
\bottomrule
\end{tabular}
\end{table}

\begin{table}[htb]
\caption{Specimen counts for lakes with samples from only one year.}
\centering
\small
\label{tab:sites_appendix2}
\begin{tabular}{llrr}
\toprule
Lake & Site & 2021 & 2022 \\
\midrule
9 Alajärvi & ala1 & 686 & 0 \\
\midrule
\multirow[t]{3}{*}{10 Hauhonselka} & hau1 & 375 & 0 \\
 & hau2 & 321 & 0 \\
 & hau3 & 503 & 0 \\
\midrule
\multirow[t]{2}{*}{11 Kajoonjärvi} & kaj1 & 233 & 0 \\
 & kaj2 & 412 & 0 \\
\midrule
\multirow[t]{3}{*}{12 Kakskerranjärvi} & kak1 & 313 & 0 \\
 & kak2 & 388 & 0 \\
 & kak3 & 455 & 0 \\
\midrule
\multirow[t]{2}{*}{13 Köylionjärvi} & koy1 & 648 & 0 \\
 & koy2 & 742 & 0 \\
\midrule
\multirow[t]{2}{*}{14 Kuhajärvi} & kuh1 & 385 & 0 \\
 & kuh2 & 555 & 0 \\
\midrule
\multirow[t]{3}{*}{15 Lopen Pääjärvi} & lop1 & 372 & 0 \\
 & lop2 & 174 & 0 \\
 & lop3 & 344 & 0 \\
\midrule
\multirow[t]{3}{*}{16 Siika-Kämä} & sii1 & 1271 & 0 \\
 & sii2 & 277 & 0 \\
 & sii3 & 875 & 0 \\
\midrule
\multirow[t]{2}{*}{17 Viitaanjärvi} & vii1 & 205 & 0 \\
 & vii2 & 132 & 0 \\
\midrule
\multirow[t]{3}{*}{18 Hiidenvesi} & hii1 & 0 & 337 \\
 & hii2 & 0 & 600 \\
 & hii3 & 0 & 331 \\
\midrule
19 Iso Vatjusjärvi & iva1 & 0 & 1332 \\
\midrule
\multirow[t]{3}{*}{20 Komujärvi} & kom1 & 0 & 801 \\
 & kom2 & 0 & 1542 \\
 & kom3 & 0 & 1091 \\
\midrule
21 Ullavanjärvi & ull1 & 0 & 1912 \\
\midrule
22 Viekijärvi & vie1 & 0 & 1044 \\
\bottomrule
\end{tabular}
\end{table}

\begin{figure*}
    \centering
    \includegraphics[width=0.8\linewidth]{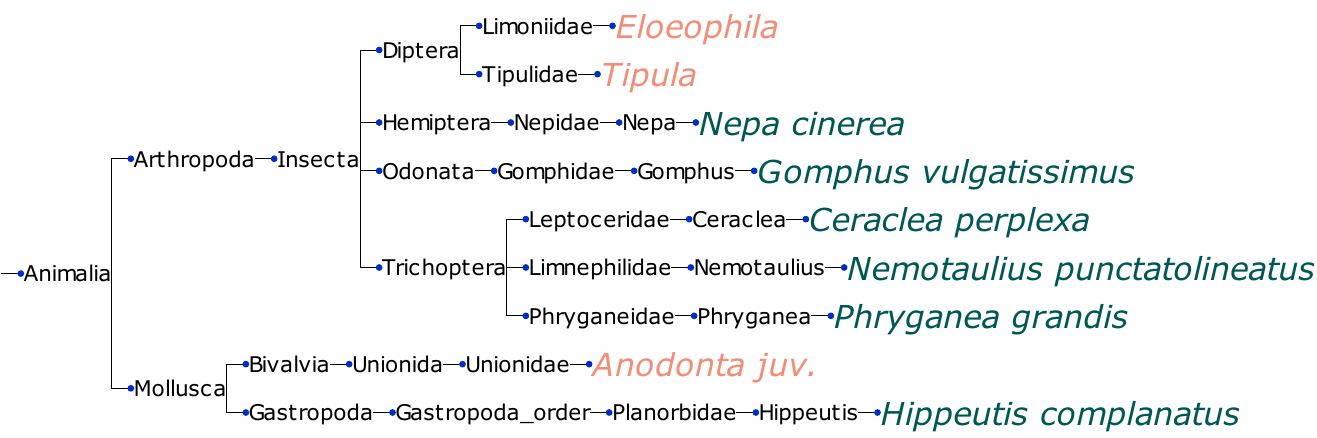}
    \caption{Taxa that were collected during the monitoring but we were not able to take images of, due to the specimens being too large or too small to be imaged.}
    \label{fig:missing-taxa}
\end{figure*}

\begin{figure*}[htb]
    \centering
    \includegraphics[width=0.9\linewidth]{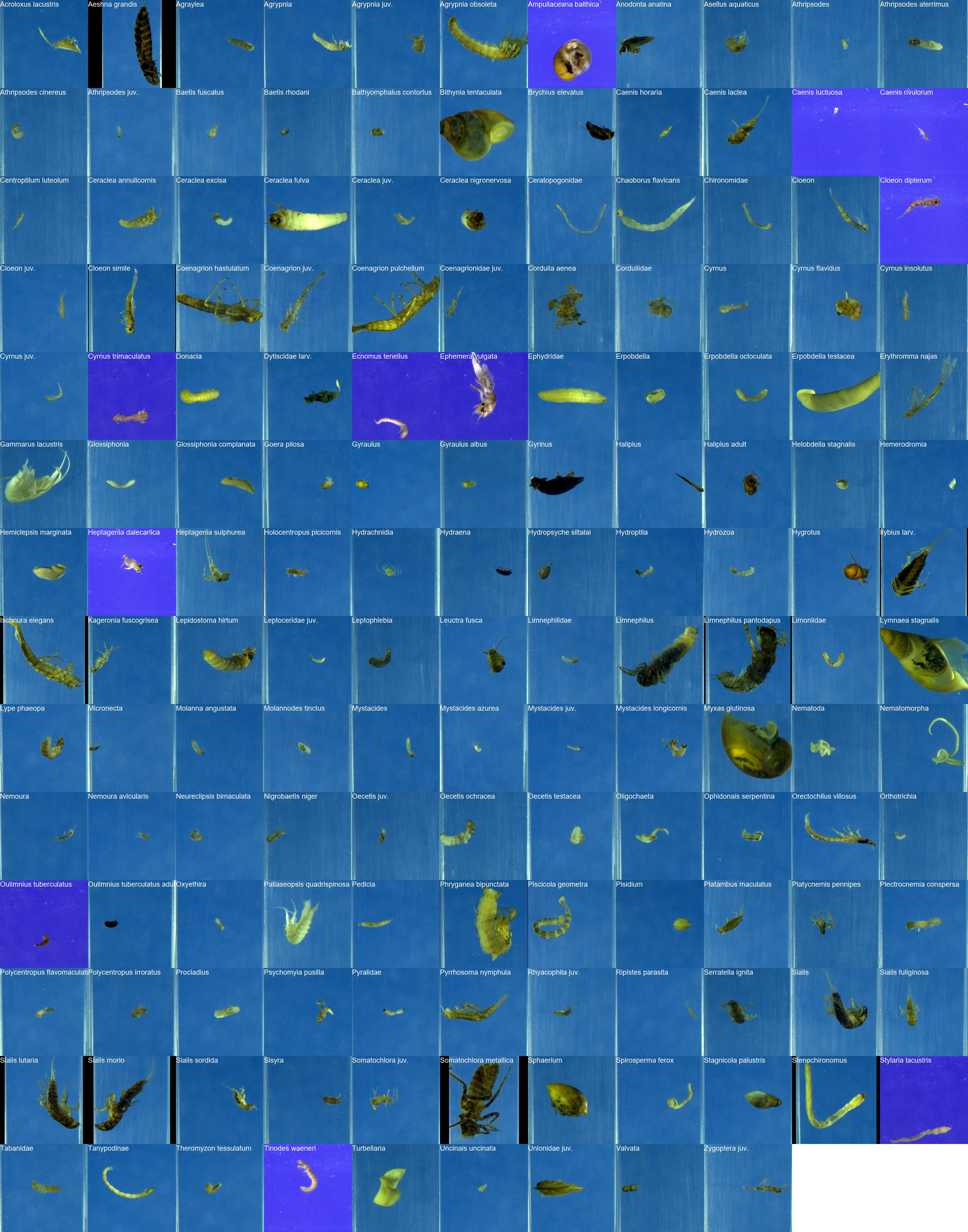}
    \caption{A randomly chosen example thumbnail image from each of the 152 classes. The images have slight brightness and contrast adjustments for illustrative purposes.}
    \label{fig:image-example-all}
\end{figure*}

\subsection{Splits}
\label{app:splits}

The classification and few-shot benchmarks are split to five cross-validation folds, where the test sets are mutually exclusive. This allows for jackknife-style cross-validation techniques where the test sets are aggregated together before final evaluations.
Image and specimen counts for first fold for classification and few-shot tasks, used in the main paper experiments, are given in Table~\ref{tab:train-test-splits}.

The classes in the monitor task train and test splits are non-overlapping. The dataset has 152 classes in total. The train split (2021 specimens) has 128 classes, and the test split (2022 specimens) has 109 classes. 85 of these classes are common for both years. The train split has 43 classes not present in the test set, and the test set has 24 classes not present in the train set. These 24 classes are also the out-of-distribution detection classes. In total there are 72 individuals and 2883 images from these classes. The difference in the sets of classes is illustrated in the monitoring benchmark confusion matrix in Fig.~\ref{fig:confusion-monitor}.

\begin{table}[htb]
\centering
\small
\caption{Train-test-splits statistics for the first fold in the Classification and Few-shot benchmarks.}
\label{tab:train-test-splits}
\begin{tabular}{lrrr}
\toprule
         & Train & Val & Test \\
\midrule
\multicolumn{4}{c}{Classification}\\
\midrule
Images & 1,882,046                   & 282,923                  & 543,054                   \\
\hspace{3mm}\textit{With DNA}  & \textit{215,318}                    & \textit{29,638}                   & \textit{62,304}                    \\
\hspace{3mm}\textit{With biomass}  & \textit{85,256}                     & \textit{13,427}                   & \textit{21,944}                    \\
        Specimens & 29,346                     & 4386                    & 8433                     \\
\hspace{3mm}\textit{With DNA} & \textit{953}                      & \textit{133}                    & \textit{263}                      \\
\hspace{3mm}\textit{With biomass} & \textit{1049}                      & \textit{157}                     & \textit{288}                    \\
\midrule
\multicolumn{4}{c}{Few-shot}\\
\midrule
Images & 30,575                   & 4426                  & 8382                   \\
\hspace{3mm}\textit{With DNA}  & \textit{487}                    & \textit{60}                   & \textit{145}                    \\
\hspace{3mm}\textit{With biomass}  & \textit{0}                     & \textit{0}                   & \textit{0}                    \\
        Specimens & 621                     & 93                    & 179                     \\
\hspace{3mm}\textit{With DNA} & \textit{8}                      & \textit{1}                    & \textit{2}                      \\
\hspace{3mm}\textit{With biomass} & \textit{0}                      & \textit{0}                     & \textit{0}                    \\

\bottomrule
\end{tabular}
\end{table}

\clearpage

\begin{figure}[htb]
    \centering
    \includegraphics[width=\linewidth]{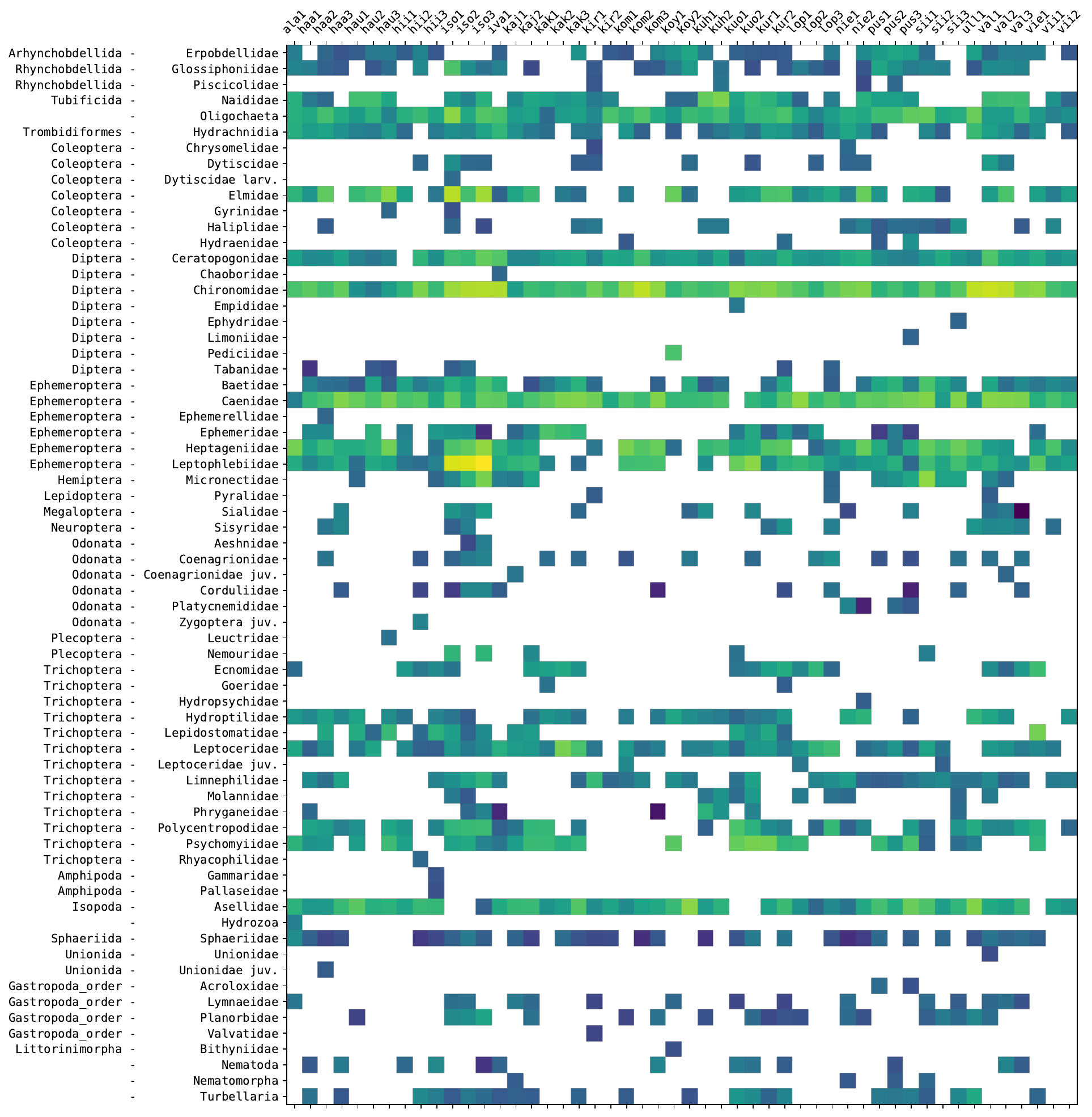}
    \caption{Heatmap representing the number of images from each taxonomic family and sampling site. Brighter yellow corresponds to a higher number of images on a logarithmic scale.}
    \label{fig:heatmap}
\end{figure}

\clearpage

\subsection{Sampling}
\label{app:sampling}

Each lake has 1-3 sampling sites from rocky shore areas with a water depth of 25-40cm. The sites are from different areas of the same lake. The sampling is performed with a kick-sampling process defined in the EU Water Framework Directive \cite{european2000directive}. During kick-sampling, the lake bed is disturbed by kicking repeatedly and released material is collected with a net. Two, three, or six 20 second kick-samples are collected from each site, depending on the number of suitable sampling sites in the lake, so that each lake has a total of six samples.

\subsubsection{Dataset coverage analysis}
\label{app:coverage}

We received 1013 unique containers for taxon+sampling site pairs for each year (total of 2026; the equal number is by chance). We were able to image specimens from all but 30 containers from the year 2021, and all but 17 containers from the year 2021. The containers we did not image either contained no specimens or had specimens that were too large to be imaged. The nine missing taxa are given in Fig.~\ref{fig:missing-taxa}.

As some specimens were not imaged, the set of images does not always statistically represent the monitoring program specimen counts per sampling site.
To find how well the imaged data reflects the original monitoring, we performed bootstrap sampling with replacement for to estimate the confidence interval [2.5\%, 97.5\%] for each site-taxon count. If all taxa for a site were imaged and their imaged counts fall within this interval, we consider the site to be well-represented.

\begin{figure*}[ht]
    \centering
\includegraphics[width=\linewidth]{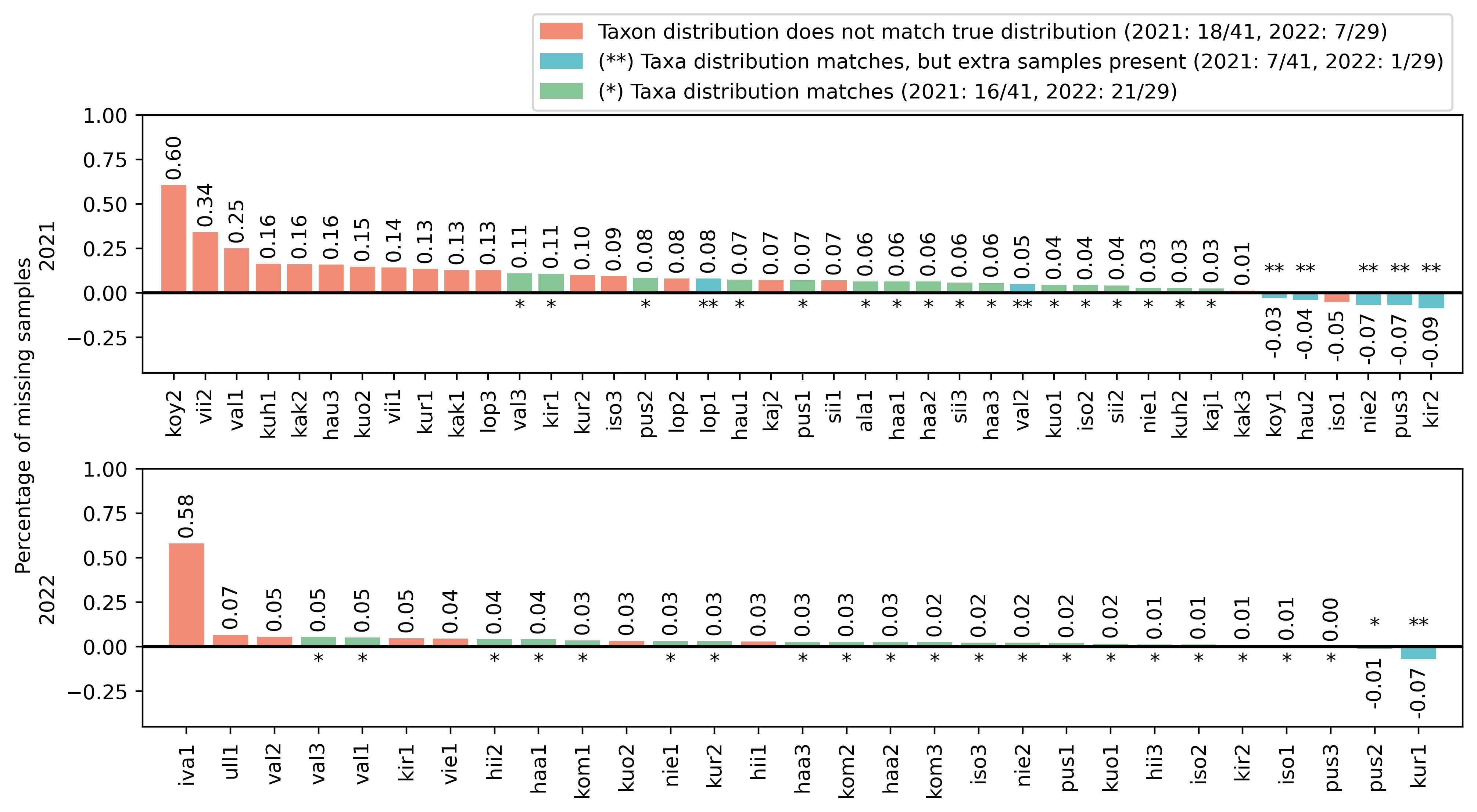}
    \caption{Percentage of specimens per well-represented sampling site we were not able to image. }
    \label{fig:coverage}
\end{figure*}

Using this approach, we found 16/41 sites in 2021 and 21/29 sites in 2022 to be well-represented. In some cases, the number of specimens in containers was larger than reported. If we allow extra samples per site, the number of representative sites grows to 23 (2021) and 22 (2022). Nine of these well-represented sites are present in both years (12 with extra samples allowed), and meaningful comparisons that require accurate distribution information between them can be made. The sites are illustrated in Fig.~\ref{fig:coverage}.

Importantly, all available specimens were imaged without any selection or filtering, ensuring that the dataset remains unbiased, even though the image sets of all sites do not strictly match the monitoring program counts.
The representativeness analysis can be used to find sites where quantitative analyses (e.g., comparing size distributions and trait diversity) can be interpreted with highter statistical confidence.

\subsection{Biomass and size measurements}
\label{app:biomass}
The biomass subset was chosen to include specimens from taxa that had over 20 specimens and were morphologically identified to the species level. Due to time constraints, we estimated to be able to process around 1500 specimens. The final specimen count was chosen to represent the true distribution across taxa. This means that the distribution is imbalanced as in real life. We made this design choice to accommodate simulating real-life sampling and metabarcoding scenarios. The specimen sampling was stratified by sampling site, meaning that all sampling sites are equally represented in this subset.

The specimens were weighed using aluminium weighing dishes. The dishes were first dried for 22-24 hours in 105°C and weighed empty. The specimens were set on the dishes, dried again for 22-24 hours in 105°C, and weighed with the protocol described in the main paper. All weights for both measurements are provided in the dataset metadata.

\begin{figure*}[tb]
    \centering
\includegraphics[width=\linewidth]{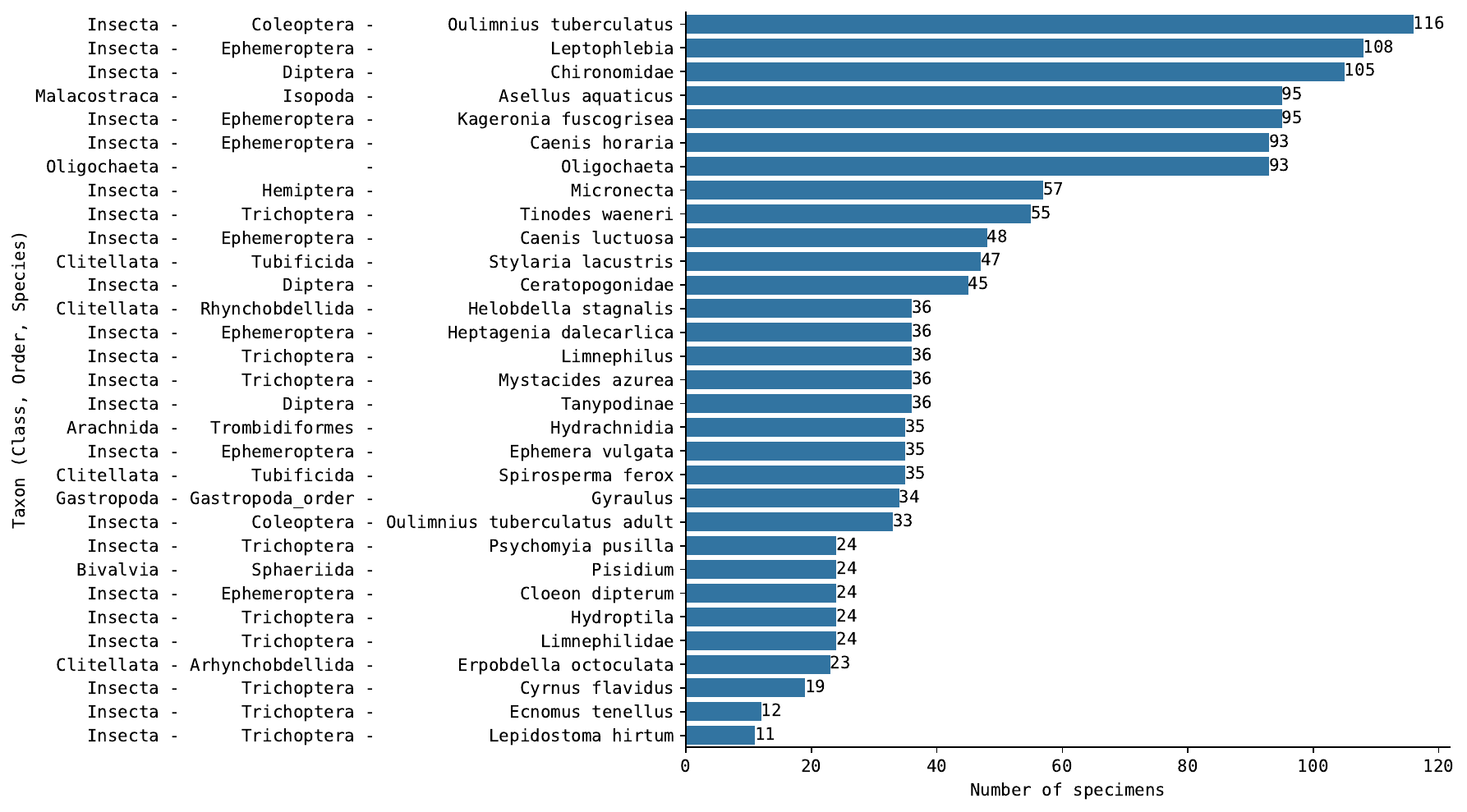}
    \caption{Number of specimens with successfully measured biomass. Taxon class, order, and species names are presented.}
    \label{fig:biomass}
    \vspace{10pt}
\end{figure*}

Fig.~\ref{fig:biomass} gives the number of specimens we were able to measure biomass from. We processed and imaged 1514 specimens in total and obtained dry mass measurements from 1494 specimens. A common failure reason for biomass measurements was the specimen being destroyed after drying and before measurement. This happened often due to static electricity causing the specimen to "jump out" of the aluminum dish.

\begin{figure*}[ht]
    \centering
    \begin{subfigure}[]{0.5\textwidth}
        \includegraphics[width=\textwidth]{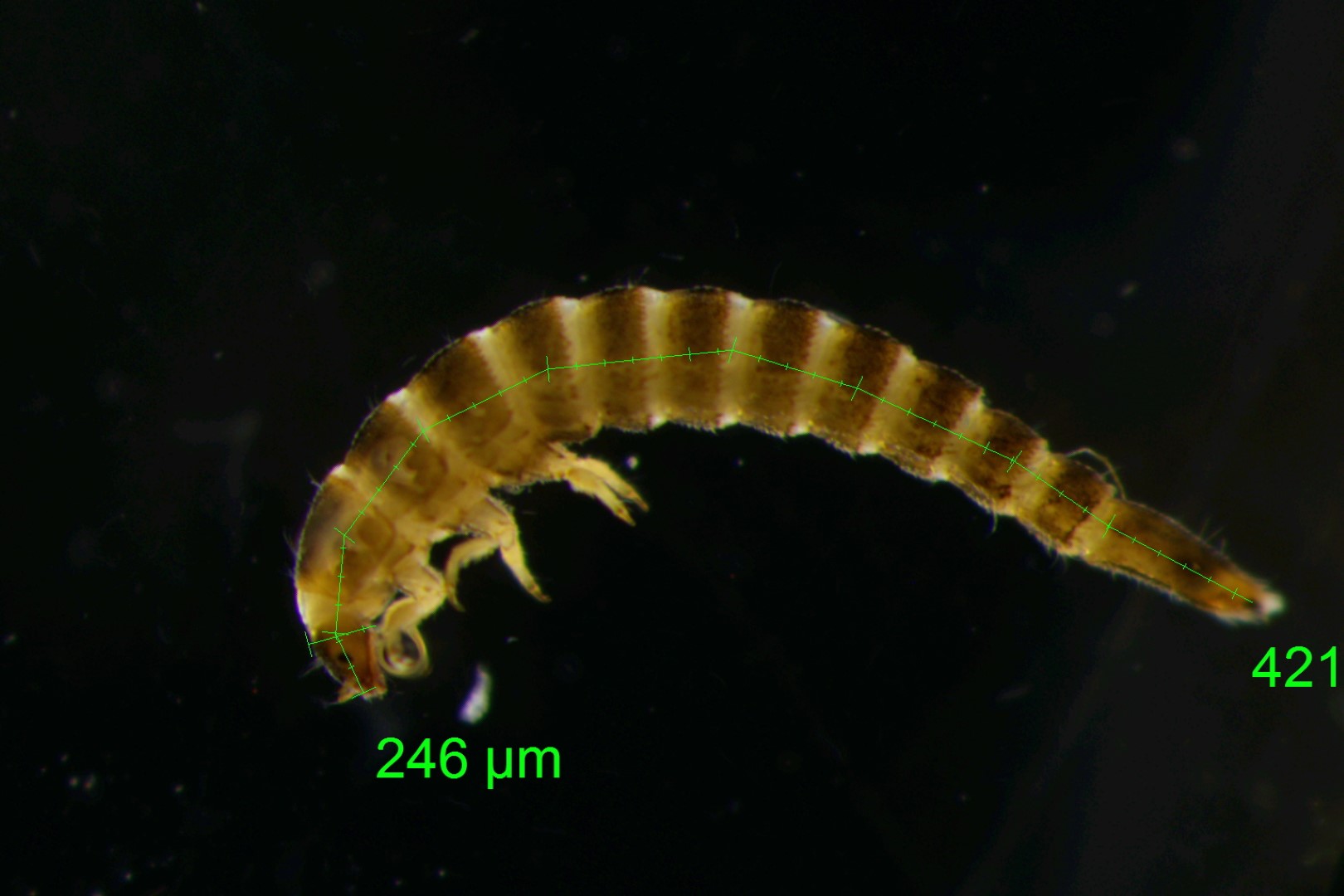}
        \caption{\textit{Ouliminius tuberculatus}}
    \end{subfigure}\hfill
    \begin{subfigure}[]{0.5\textwidth}
        \includegraphics[width=\textwidth]{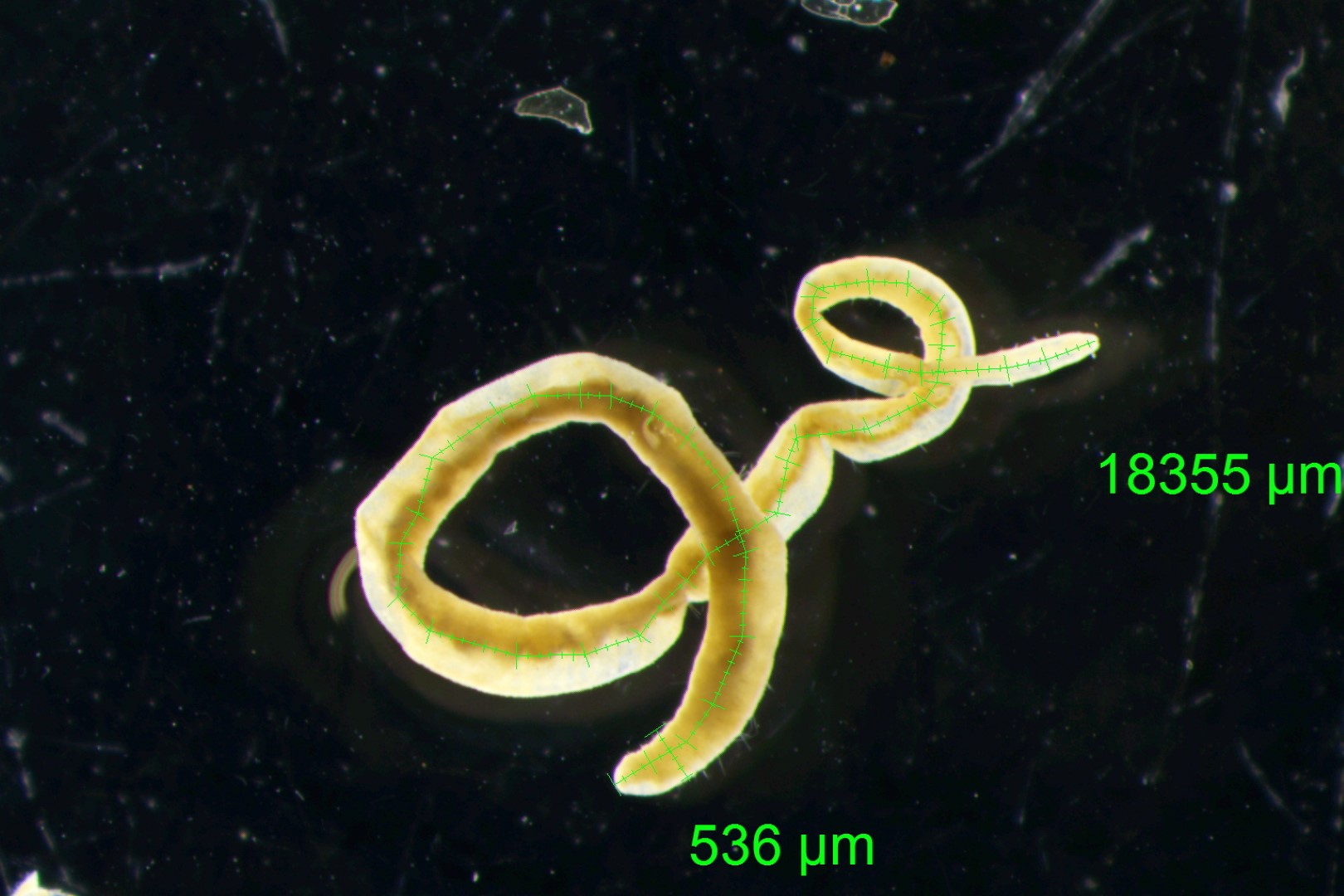}
        \caption{\textit{Spirosperma ferox}}
    \end{subfigure}\\
    \begin{subfigure}[]{0.5\textwidth}
        \includegraphics[width=\textwidth]{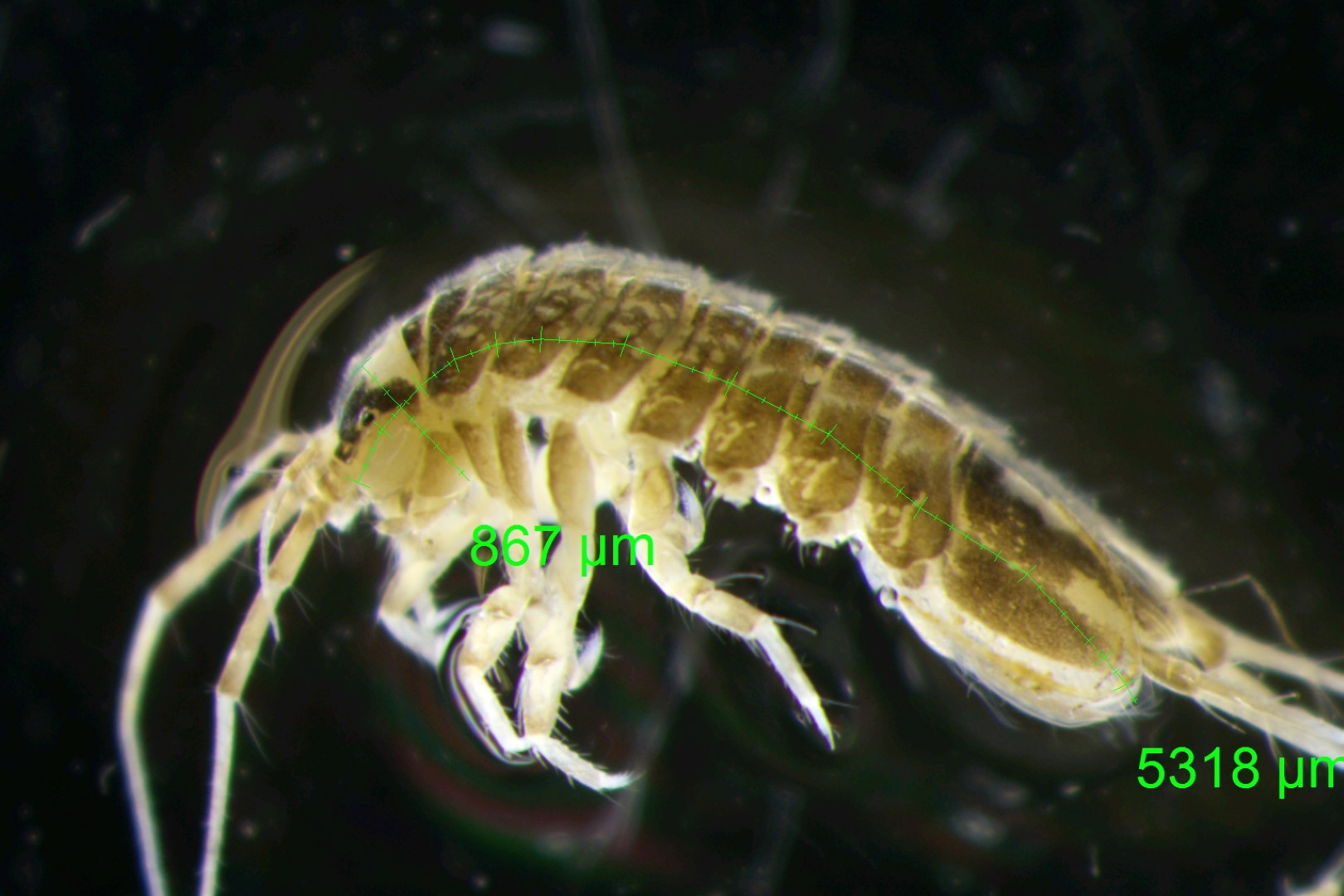}
        \caption{\textit{Asellus aquaticus}}
    \end{subfigure}\hfill
    \begin{subfigure}[]{0.5\textwidth}
        \includegraphics[width=\textwidth]{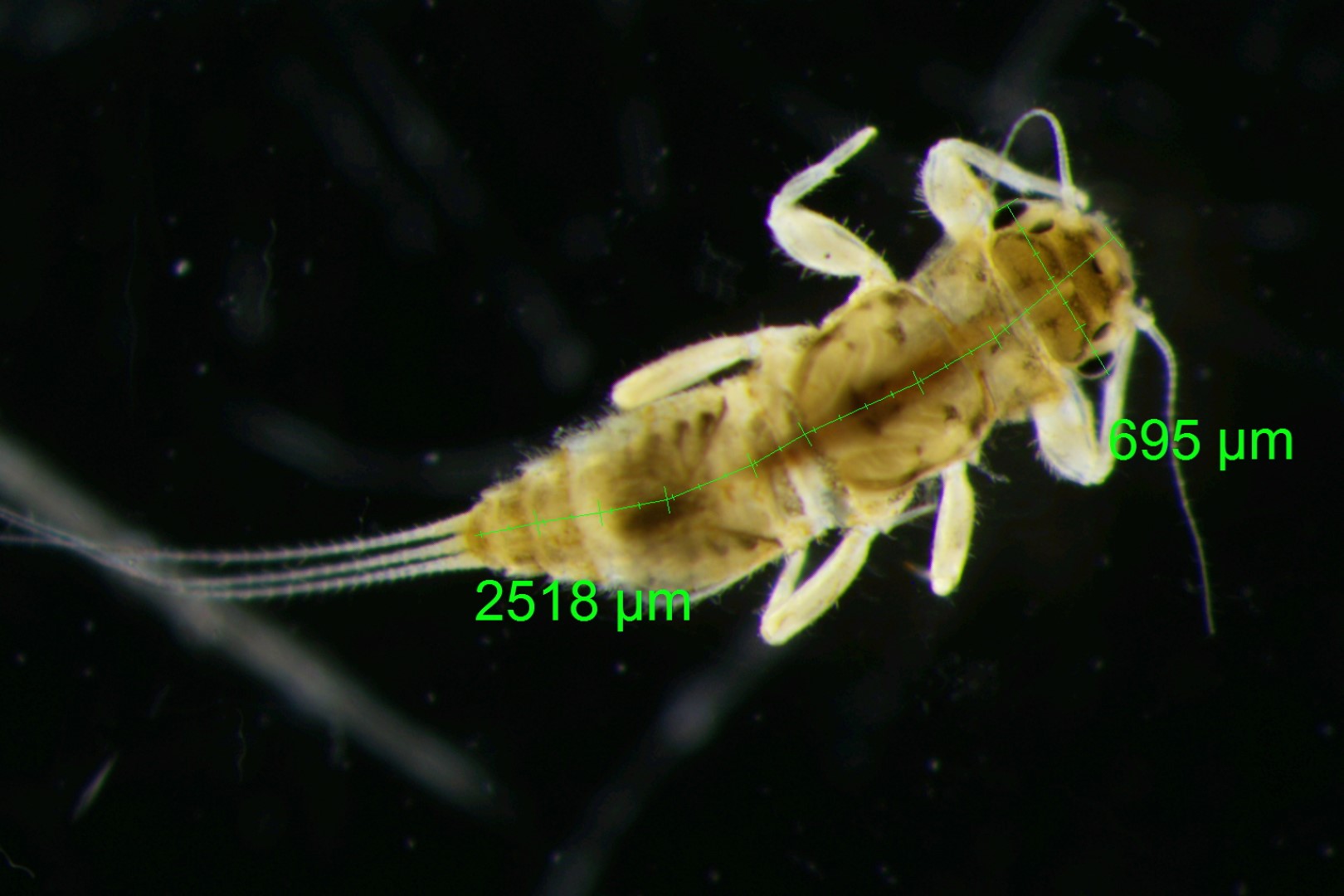}
        \caption{\textit{Caenis luctuosa}}
    \end{subfigure}
    \caption{Examples of high-resolution images and their measurements in the biomass subset}
    \label{fig:hi-res}
\end{figure*}

The specimens were measured using DeltaPix InSight microscope software suite (version 6.6.2). Because the specimens in the biomass subset were imaged with the microscope during measurements, we have high-resolution images of these specimens in addition to the dual-view image sequences. Examples of these high-resolution images and the measurements are in Fig.~\ref{fig:hi-res}.

\subsection{DNA subset collection}
\label{app:dna}

\begin{figure*}[htb]
    \centering
\includegraphics[width=\linewidth]{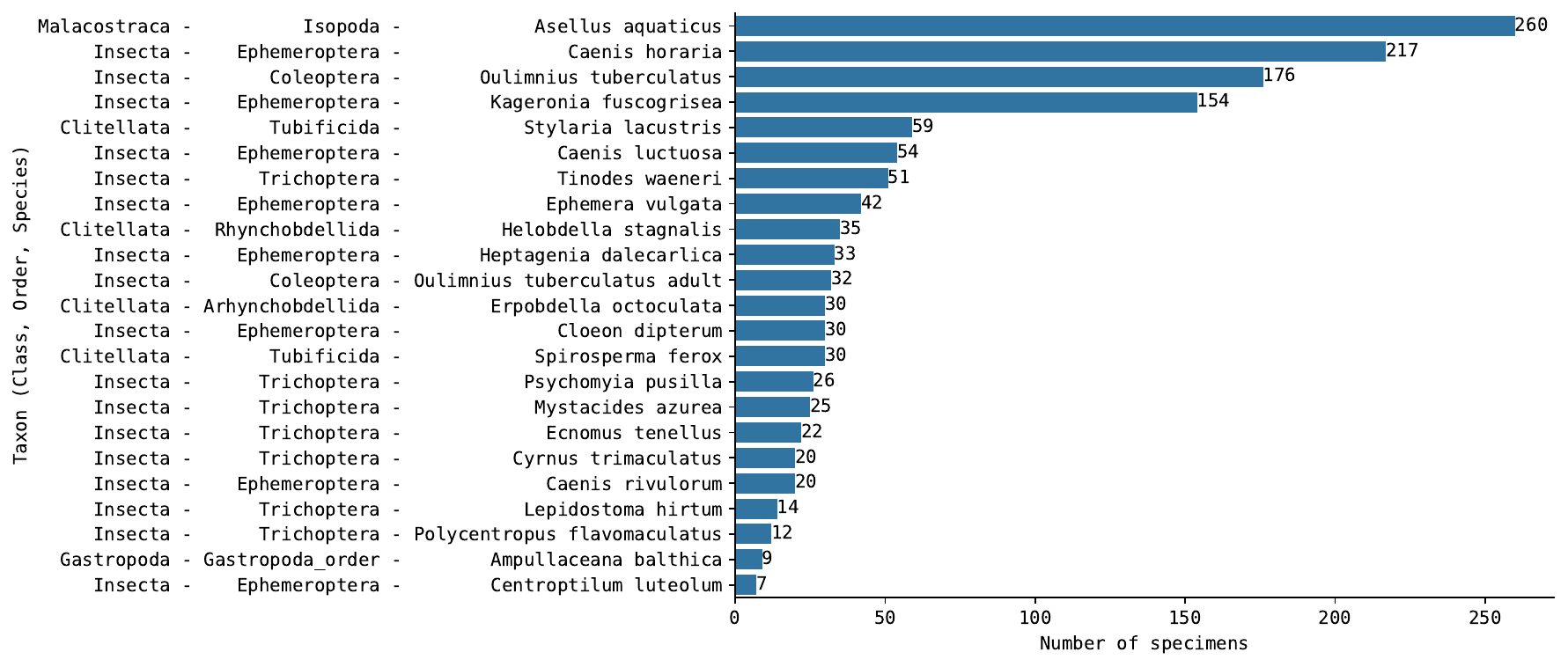}
    \caption{Number of specimens with successfully sequenced DNA. Taxon class, order and species names are presented.}
    \label{fig:barcoding}
\end{figure*}

The DNA subset was chosen with similar criteria as the biomass specimens. These samples were also stratified by sampling site. Fig.~\ref{fig:barcoding} gives the number of specimens from each taxa we obtained DNA from. We processed and imaged in total 1518 specimens, but due to the nature of DNA extraction and sequencing, we did not obtain DNA for all specimens.

\subsubsection{DNA extraction}
DNA was extracted using 100 $\mu$L of 5\% Chelex solution (BioRad). Depending on the size of the specimen, either 2-4 legs were manually detached from the specimen or the complete specimen was used. After that, samples were incubated for 20 minutes at 96°C.

\subsubsection{PCR}
A two-step polymerase chain reaction (PCR) approach was used to amplify the target COI gene fragment. In the first step PCR, tagged fwhF2 and fwhR2n \cite{vamos2017Short} primers were used to amplify a 205 bp fragment of the COI gene in 10 $\mu$L reactions. 5 $\mu$L DreamTaq master mix (Thermo Fisher Scientific), 0.2 $\mu$L forward primer (fwhF2, 10 $\mu$M), 0.2 $\mu$L reverse primer (fwhR2n, 10 $\mu$M), 3.6 $\mu$L H2O, and 1 $\mu$L DNA extract per reaction were used. The PCR started with 3 min of initial denaturation at 95°C, followed by 40 cycles of 30 s of denaturation at 95°C, 30 s of primer annealing at 58°C, and 1 min of elongation at 72°C, and a final amplification for 15 min at 72°C. During the second step PCR, additionally tagged Illumina primers were used to prepare the amplicons for sequencing in 10 $\mu$L reactions 5 $\mu$L DreamTaq master mix (Thermo Fisher Scientific), 2 $\mu$L tagging primers (100 $\mu$M), 1 $\mu$L H2O, and 2 $\mu$L DNA product from the first PCR per reaction were used. The PCR started with 5 min of initial denaturation at 95°C, followed by 10 cycles of 30 s of denaturation at 95°C, 1 min 30 s of primer annealing at 61°C, and 30 s of elongation at 72°C, and a final amplification for 10 min at 68°C.

\subsubsection{Library preparation}
To prepare the samples for sequencing, additional cleaning steps were performed. First, all samples were pooled, and GuHCl-buffer was added in a ratio of 2:1 (14 mL GuHCl and 7 mL library). The mixture was run through a 30 mL silica gel column with a vacuum manifold. Two times 10 mL of wash buffer were added and run through the column subsequently. To dry the column, it was centrifuged for 2750 × g for one minute. 1000 $\mu$L of elution buffer was added and after a 3-minute incubation period, the tube was centrifuged at 2750 × g again. This process was repeated with a smaller silica gel column, with a starting volume of 1 mL. 2 mL of GuHCl-buffer were added to the eluate and run through the silica column, followed by adding 650 $\mu$L wash buffer twice. Finally, the DNA was eluted by adding 100 $\mu$L of elution buffer, incubating for one minute and centrifuging.
The DNA concentration of the library was checked with Qubit (Thermo Fisher Scientific) measurements following the manufacturers’ instructions.

To remove potential bubble products that have formed during PCR, a reconditioning PCR was performed. For this, a reconditioning PCR was prepared with 225 $\mu$L TaqManTM Multiplex Master Mix (Thermo Fisher Scientific), 2.25 $\mu$L Illumina P5 and P7 primer (100 $\mu$M) each, 19 $\mu$L H2O and 202 $\mu$L template and split into four reactions (single cycle protocol: Denaturation for 5.5 min at 95°C, annealing for 1.5 min at 60°C and extension for 1.5 min at 72°C, followed by a final extension for 10 min at 68°C). After pooling the products of the four PCRs together, an additional clean up (see above) was performed.
To remove any leftover DNA-fragments, such as primers and nuclear DNA, a bead-based size selection was performed. For this, 70 $\mu$L of clean-up beads were added to 100 $\mu$L of the cleaned-up eluate in a 1.5 mL Eppendorf tube and incubated for five minutes to bind the DNA to the beads. After 2 min of incubation on a custom-made magnetic rack, the supernatant was discarded. Two additional washing steps were performed by covering the beads with 500 $\mu$L of wash buffer, 30 sec of incubation and discarding the supernatant. After this, the beads were dried for five minutes. To elute the DNA from the magnetic beads, 50 $\mu$L elution buffer was added and incubated for 5 min. The tube was then placed on the magnetic rack again and the supernatant was transferred to a new Eppendorf tube. The DNA concentration of the final library was measured with Qubit following the manufacturers’ instructions. To ensure all remaining DNA fragments matched the desired target length, a Fragment Analyzer (5200 Fragment Analyser System, Agilent) was used. Cooled samples were sent for sequencing at CeGaT (Tuebingen, Germany).

\subsubsection{Sequencing}
1632 samples (including 114 negative controls) have been sequenced on a MiSeq 300 cycle nano V2 flow cell (Read length: 2 x 150 bp, Theoretical output: 0.3 Gb (1 M clusters)) resulting in a total of 602,686 raw reads (raw data are available in FASTQ format \cite{cock2010Sanger}. Illumina index reads were demultiplexed with Illumina bcl2fastq (2.20). Adapters were trimmed with Skewer (version 0.2.2) \cite{jiang2014Skewer}. Raw data (without Illumina adapters) with quality score (phred+33 encoding) are available in FASTQ format (.fastq.gz). Per sample, two FASTQ files are given: One per read direction (forward and reverse)

\subsubsection{Bioinformatics}
Reads were further demultiplexed with Demultiplexer (version 1.1.0) since additional inline tags were combined with index reads. Tags were removed during demultiplexing and samples were saved with their respective imaging names (plate and well position and specimen ID) in FASTQ format (.fastq.gz). Paired-end merging, primer trimming, quality filtering, OTU clustering and denoising, as well as OTU filtering was done with APSCALE (version 1.6.3) \cite{buchner2022APSCALE}. OTU sequences were queried against BOLD database to assign taxonomy using BOLDigger(version 2.1.3) \cite{buchner2020BOLDigger}. 461,972 reads passed the quality filtering and were clustered into 85 OTUs.

\section{Supplementary material for Section 4 Benchmark experiments and results}

\subsection{Experimental setup}
\label{app:experimental_setup}

Tables~\ref{tab:training_details} and \ref{tab:weight_names} show more details on the models trained for the Monitoring benchmark. Most of the models and their pretrained weights are from the \texttt{timm}-library \cite{rw2019timm}, except for the BioCLIP models that are from Huggingface Hub.
The models trained for the classification task are given in Table~\ref{tab:training_details_clf}. These models are also the ones used for the few-shot task.

For the few-shot task, we used also pretrained models that were not trained further. We used three CLIP models, using weights from BioCLIP, OpenAI, and OpenCLIP \cite{ilharco_gabriel_2021_5143773}. BioCLIP and CLIP/OpenAI models use a ViT-B/16 architecture, while the OpenCLIP model uses the ViT-B/32 architecture, with LAION weights (\texttt{laion2b\_s34b\_b79k}) \cite{schuhmann2022laionb}. The DINO model is similarly a ViT-B/16, while the DINOv2 model is a ViT-B/14, both loaded from Torch Hub. The SigLIP and SigLIP2 weights were loaded using Huggingface Transformers.

All models were trained on a computing cluster provided by CSC - IT Center for Science, Finland, using four or two Tesla V100-SXM2-32GB GPUs.
Training times for all models are reported in Table~\ref{tab:training_details}.
Batch size was chosen to be the largest that can fit to the GPU memory. With four GPUs, the effective batch size is four times the size in Table~\ref{tab:training_details}..

The dataset has multi-view image sequences for each specimen, and there are various ways of using information from multiple images. Our baselines use a simple approach for fusing the information from the sequence and both views: logit averaging.
Each image frame for a specimen is classified separately using the models above. The models produce logit outputs for each class. These outputs are averaged over all frames for a single specimen. The effect of performing this fusion on different levels is provided in the main paper Table~8.

The multiview model uses two feature encoders that are similar to the single-view model. Both encoders have unique weights. Each encoder is passed an image from one of two perpendicular views of the specimen. The feature vectors from both encoders are then concatenated, and passed to a final linear projection head. Image sequences are averaged using the logit mean, similarly to the single-view case.

The multi-view model does not use all the same specimens as the single-view models for evaluation. The multi-view model is only evaluated on the specimens that have images from both cameras. This accounts for all but 222 specimens. Similarly, the ensemble model uses only these specimens.

The ensemble model uses the three best performing models based on validation dataset accuracy: Swin-T, Swin-T multiview, and EfficientNet-B4. This also combines three approaches for handling the images: single-view and multi-view models using resolution 224x224 (Swin-T, Swin-T multiview), and a single-view model using a larger resolution 320x320. The ensembling fusion is performed similarly as the sequence and multi-view fusion above by averaging the logit outputs across models.

\begin{table}[htb]
\centering
\caption{Training details of models trained on the monitoring training set.}
\label{tab:training_details}
\scriptsize
\begin{tabular}{llllll}
\toprule
Model                 & Epochs & Input size & Training time (hours) & GPUs & Batch size \\ \midrule
MobileNetV3           & 100    & 224        & 9.36                  & 4     & 256        \\
ResNet-50             & 100    & 224        & 19.56                 & 4     & 256        \\
ResNet-101            & 100    & 224        & 28.53                 & 4     & 256        \\
EfficientNet-B0       & 100    & 224        & 15.65                 & 4     & 256        \\
EfficientNet-B4       & 20     & 320        & 14.16                  & 4     & 64         \\
Swin-T                & 100    & 224        & 27.65                 & 4     & 256        \\
Swin-T (multiview)    & 100    & 224        & 24.13                 & 4     & 128        \\
Swin-B                & 100    & 224        & 57.64                 & 4     & 128        \\
ViT-B/16              & 100    & 224        & 32.26                 & 4     & 256        \\
ViT-B/16 (BioCLIP)    & 100    & 224        & 37.43                 & 4     & 256        \\
ViT-B/16 (BioCLIP-FT) & 100    & 224        & 16.51                 & 4     & 256        \\
ViT-L/14              & 100    & 224        & 133.99                & 4     & 64         \\
\bottomrule
\end{tabular}
\end{table}

\begin{table}[htb]
\centering
\caption{Training details of models trained on the classification training set. The same models are later used in the few-shot task.}
\label{tab:training_details_clf}
\scriptsize
\begin{tabular}{llllll}
\toprule
Model              & Epochs & Input size & Training time (hours) & GPUs & Batch size \\
MobileNetV3        & 100    & 224        & 11.01                 & 4     & 256        \\
ResNet50           & 100    & 224        & 26.98                 & 2     & 256        \\
EfficientNet-B0    & 100    & 224        & 19.46                 & 4     & 256        \\
EfficientNet-B4    & 20     & 320        & 16.52                 & 4     & 64         \\
Swin-T             & 100    & 224        & 36.35                 & 4     & 256        \\
Swin-T (Multiview) & 100    & 224        & 27.76                 & 4     & 128 \\
\bottomrule
\end{tabular}
\end{table}

\begin{table}[htb]
\centering
\caption{Model weight names for both fully trained models (upper section) and training-free models used for few-shot learning (lower section)}
\label{tab:weight_names}
\scriptsize
\begin{tabular}{ll}
\toprule
Model                 & Weight name \\ \midrule
MobileNetV3           & \texttt{mobilenetv3\_small\_075.lamb\_in1k} \\
ResNet-50             & \texttt{resnet50.a1\_in1k} \\
ResNet-101            & \texttt{resnet101.a1h\_in1k} \\
EfficientNet-B0       & \texttt{efficientnet\_b0.ra\_in1k} \\
EfficientNet-B4       & \texttt{efficientnet\_b4.ra2\_in1k} \\
Swin-T                & \texttt{swin\_tiny\_patch4\_window7\_224.ms\_in1k} \\
Swin-T (multiview)    & \texttt{swin\_tiny\_patch4\_window7\_224.ms\_in1k} \\
Swin-B                & \texttt{swin\_base\_patch4\_window7\_224.ms\_in22k\_ft\_in1k} \\
ViT-B/16              & \texttt{vit\_base\_patch16\_clip\_224.openai} \\
ViT-B/16 (BioCLIP)    & \texttt{hf-hub:imageomics/bioclip} \\
ViT-B/16 (BioCLIP-FT) & \texttt{hf-hub:imageomics/bioclip} \\
ViT-L/14              & \texttt{vit\_large\_patch14\_clip\_224.openai} \\
\midrule
DINO & \texttt{facebookresearch/dino:main, dino\_vitb16} \\
DINOv2 & \texttt{facebookresearch/dinov2, dinov2\_vitb14\_reg
} \\
SigLIP & \texttt{google/siglip-base-patch16-224} \\
SigLIP2 & \texttt{google/siglip2-so400m-patch16-naflex} \\
\bottomrule
\end{tabular}
\end{table}

\clearpage

\subsection{Experimental results}
\label{app:experimental_results}

The full result tables for all three benchmarks and all models are in Table~\ref{tab:monitor-fulltable} (Monitoring), Table~\ref{tab:classif-fulltable} (Classification), and Table~\ref{tab:fewshot-fulltable} (Few-shot). These tables correspond to Tables 5 and 6 in the main paper. Uncertainties are calculated by bootstrapping the final specimen-level predictions 1000 times with replacement, with $\pm$2 standard deviations reported in the tables.

\begin{figure}
    \centering
    \includegraphics[width=\linewidth]{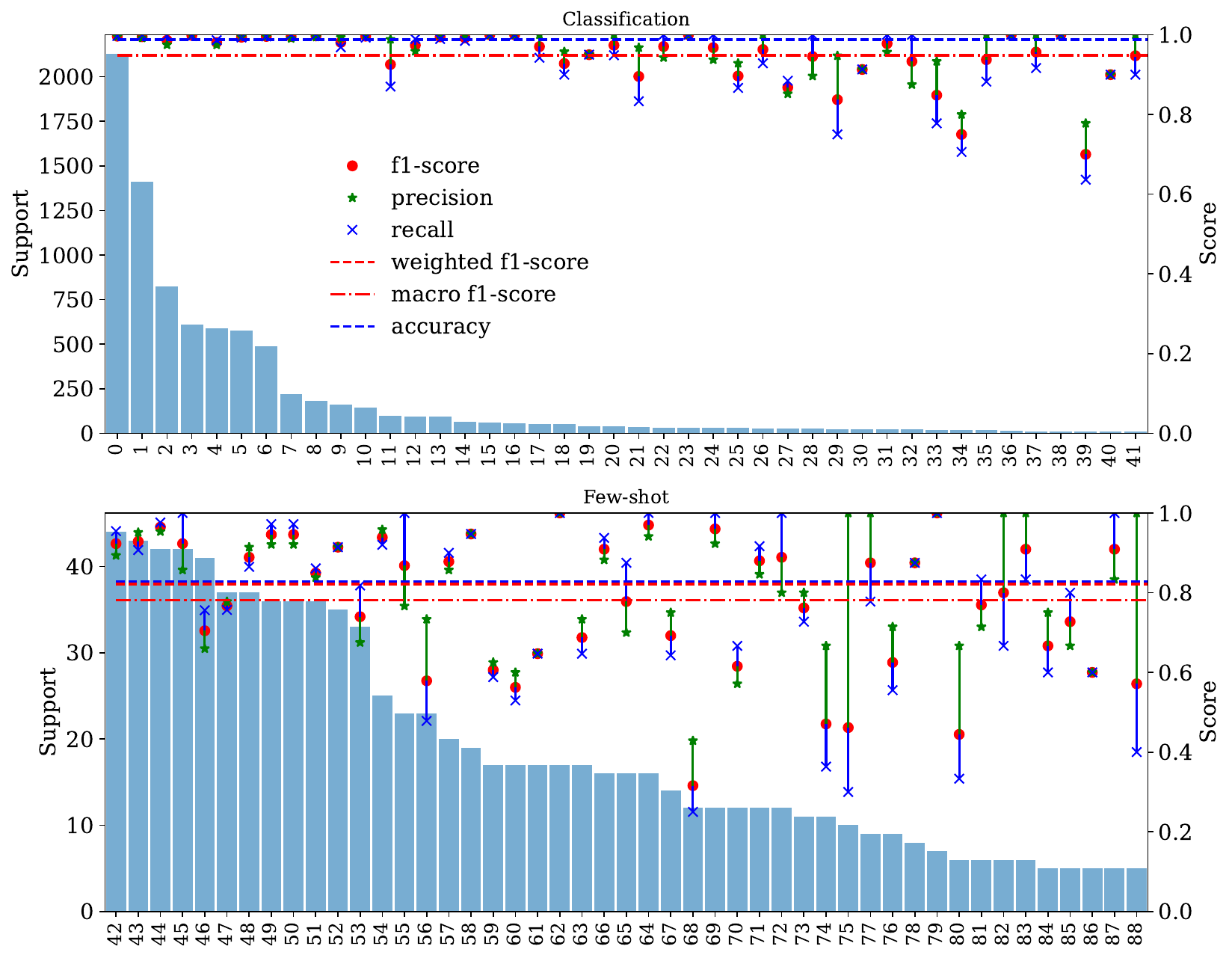}
    \caption{Classification and few-shot class-wise results. For taxon names referenced by numbers, see Table~\ref{tab:classif-fulltaxa} and Table~\ref{tab:fewshot-fulltaxa}}
    \label{fig:clf_fewshot_classwise}
\end{figure}

We report class-wise results for the best performing models. For monitoring and classification task this is the ensemble model. The Swin-T model performed the best in the few-shot task. A numerical table of the class-wise results corresponding to the monitoring task Fig. 3 in the main paper is given in Table~\ref{tab:monitor-fulltaxa}. Similar figures for the classification and few-shot tasks can be seen in Fig.~\ref{fig:clf_fewshot_classwise} and the corresponding numerical values in Table~\ref{tab:classif-fulltaxa} and Table~\ref{tab:fewshot-fulltaxa}. The number indices in the figures correspond to the numbers in the tables. The legend in Figure~\ref{fig:clf_fewshot_classwise} can be applied to Figure 3 in the main paper.

\begin{figure*}[t]
    \centering
\includegraphics[width=\linewidth]{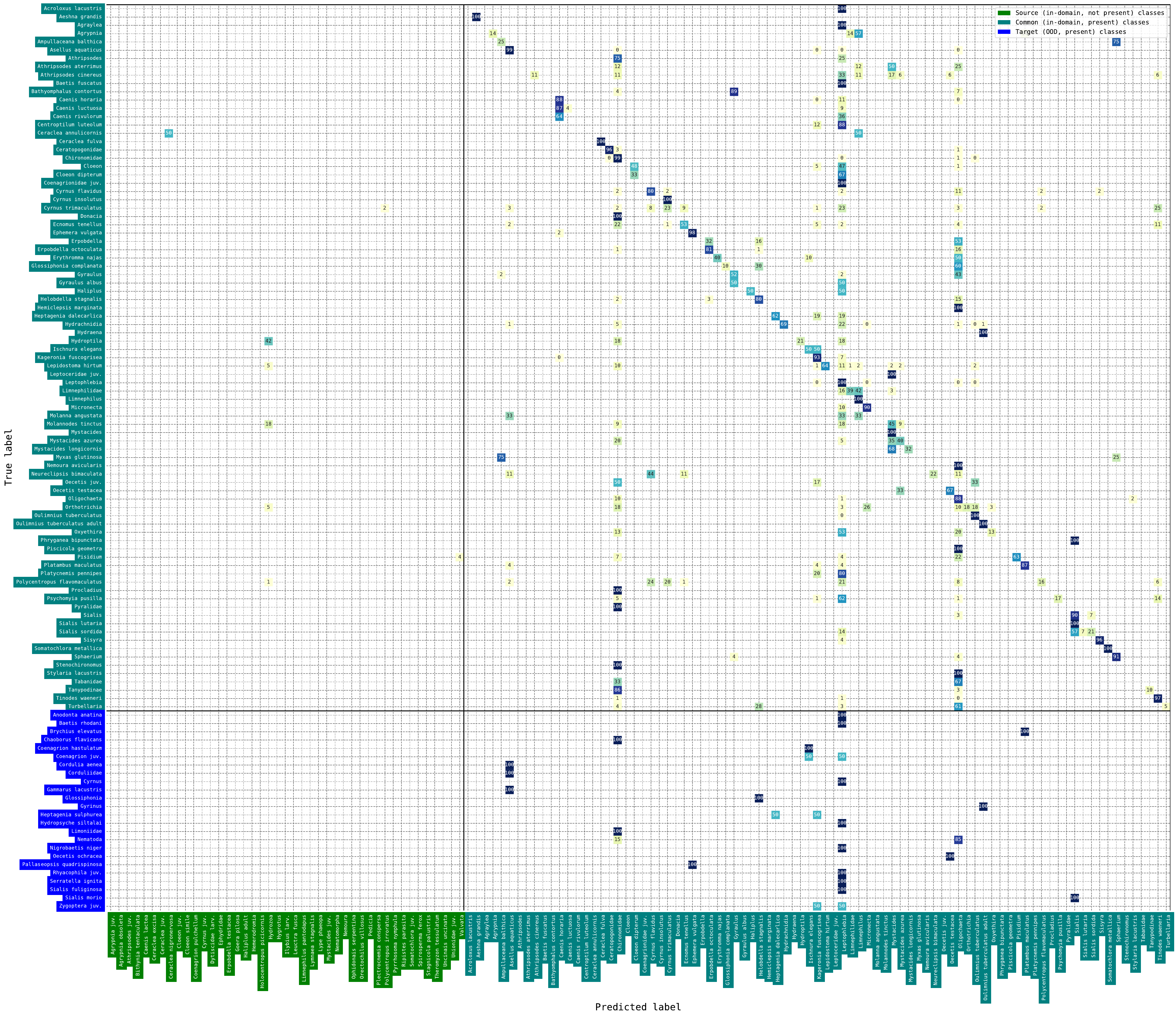}
    \caption{Full confusion matrix for the monitoring task, containing also outlier mistakes. Predictions are always made in the set of labels present in the 2021 dataset. Teal classes are the 85 common classes for train and test sets. Green classes are classes unique to the train set. Blue classes are the OOD classes unique to the test set. Values are percentages of true values and rows sum to one hundred.}
    \label{fig:confusion-monitor}
\end{figure*}

\begin{figure*}[t]
    \centering
\includegraphics[width=\linewidth]{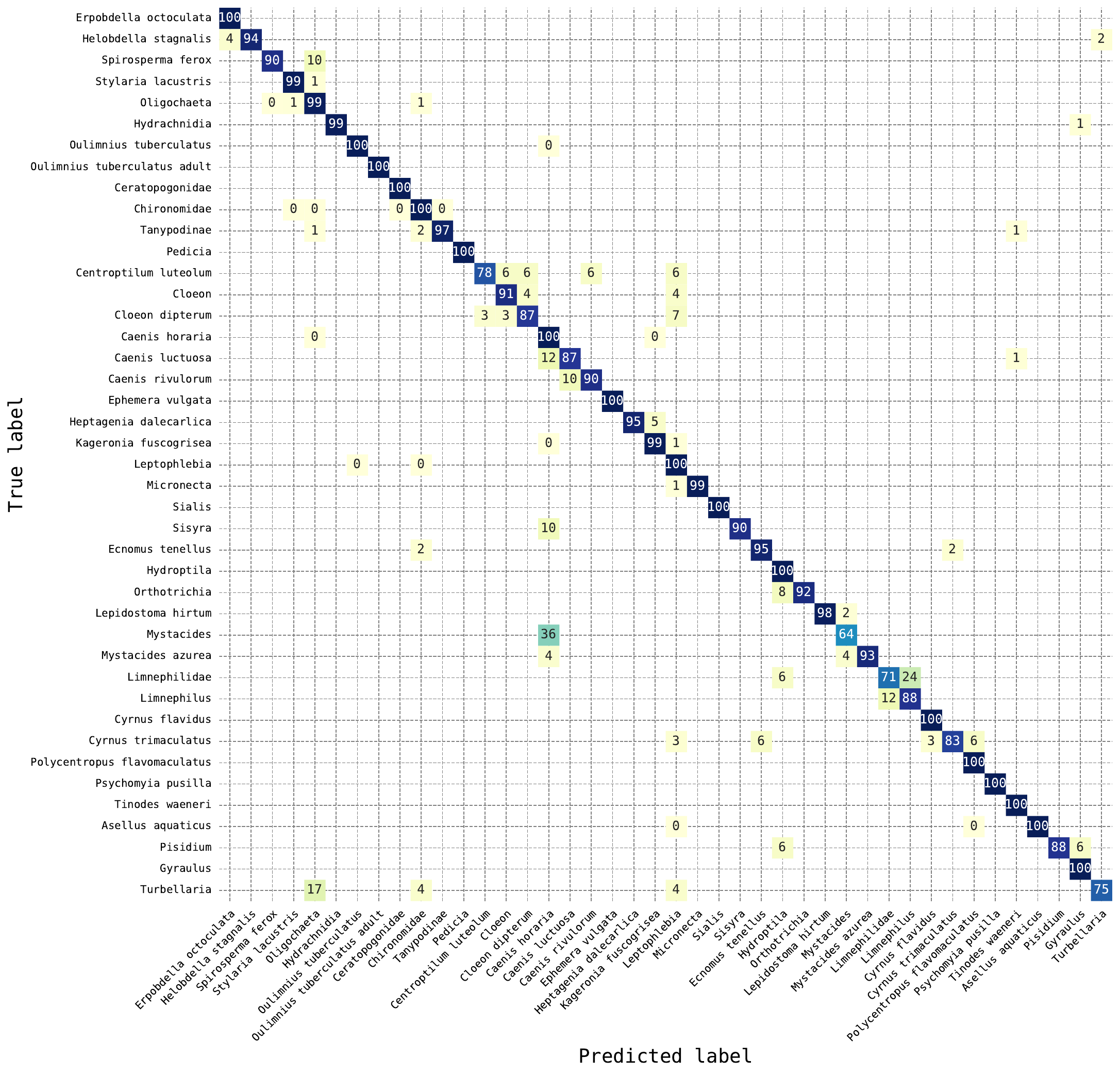}
    \caption{Standard classification confusion matrix. Values are percentages of true values and rows sum to one hundred.}
    \label{fig:confusion}
\end{figure*}

\begin{figure*}[t]
    \centering
\includegraphics[width=\linewidth]{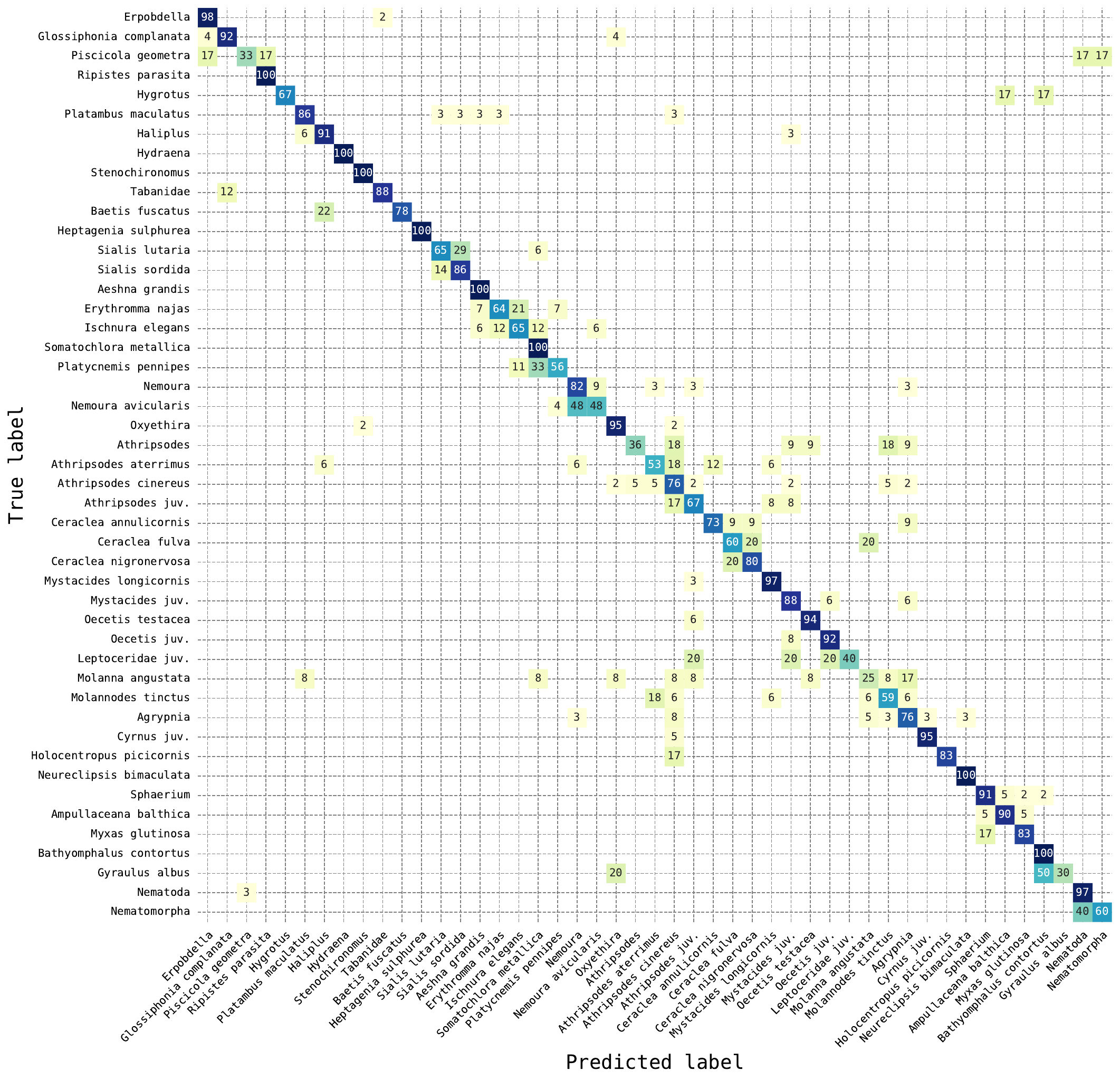}
    \caption{Few-shot classification confusion matrix. Values are percentages of true values and rows sum to one hundred.}
    \label{fig:confusion-fewshot}
\end{figure*}

Confusion matrices for these tasks are given in Fig.~\ref{fig:confusion-monitor} (Monitoring), Fig.~\ref{fig:confusion} (Classification), and Fig.~\ref{fig:confusion-fewshot} (Few-shot). The classification and few-shot confusion matrices are standard confusion matrices, normalized along the true label. The monitoring confusion matrix presents predictions across mismatching source and target label sets. The true label set is the target labeling (taxa present in 2022 dataset), and the predicted label set is the source labeling (taxa present in 2021 dataset). Because the dataset is trained with only 2021 data, it will produce predictions from this set also for out-of-distribution classes.

\input{full_table}

\clearpage

\begin{longtable}{llrrrr}
\caption{Full class-wise monitoring task results corresponding to Fig. 3 in the main paper. Index corresponds to the x-axis label in the main paper figure. Support is the number of true examples in the test set.}
\label{tab:monitor-fulltaxa}\\
\toprule
Index & Taxon & Precision & Recall & F1-score & Support \\
\midrule
0 & Chironomidae & 0.912 & 0.993 & 0.951 & 6858 \\
1 & Leptophlebia & 0.737 & 0.996 & 0.847 & 1855 \\
2 & Caenis horaria & 0.938 & 0.884 & 0.911 & 1726 \\
3 & Asellus aquaticus & 0.989 & 0.994 & 0.992 & 1248 \\
4 & Oligochaeta & 0.807 & 0.880 & 0.842 & 1119 \\
5 & Kageronia fuscogrisea & 0.969 & 0.934 & 0.951 & 1092 \\
6 & Oulimnius tuberculatus & 0.970 & 0.995 & 0.982 & 642 \\
7 & Ceratopogonidae & 0.998 & 0.958 & 0.978 & 479 \\
8 & Tanypodinae & 1.000 & 0.104 & 0.188 & 472 \\
9 & Tinodes waeneri & 0.859 & 0.975 & 0.913 & 399 \\
10 & Hydrachnidia & 1.000 & 0.691 & 0.817 & 275 \\
11 & Lepidostoma hirtum & 1.000 & 0.641 & 0.781 & 256 \\
12 & Ecnomus tenellus & 0.831 & 0.527 & 0.645 & 131 \\
13 & Cyrnus trimaculatus & 0.600 & 0.234 & 0.337 & 128 \\
14 & Oulimnius tuberculatus adult & 0.878 & 1.000 & 0.935 & 115 \\
15 & Caenis luctuosa & 1.000 & 0.043 & 0.082 & 94 \\
16 & Cloeon & 0.000 & 0.000 & 0.000 & 86 \\
17 & Polycentropus flavomaculatus & 0.824 & 0.165 & 0.275 & 85 \\
18 & Psychomyia pusilla & 1.000 & 0.167 & 0.286 & 78 \\
19 & Turbellaria & 1.000 & 0.053 & 0.100 & 76 \\
20 & Erpobdella octoculata & 0.873 & 0.809 & 0.840 & 68 \\
21 & Helobdella stagnalis & 0.650 & 0.800 & 0.717 & 65 \\
22 & Cyrnus flavidus & 0.590 & 0.803 & 0.681 & 61 \\
23 & Centroptilum luteolum & 0.000 & 0.000 & 0.000 & 50 \\
24 & Cloeon dipterum & 0.268 & 0.333 & 0.297 & 45 \\
25 & Ephemera vulgata & 1.000 & 0.977 & 0.989 & 44 \\
26 & Gyraulus & 0.460 & 0.523 & 0.489 & 44 \\
27 & Orthotrichia & 1.000 & 0.179 & 0.304 & 39 \\
28 & Hydroptila & 1.000 & 0.211 & 0.348 & 38 \\
29 & Limnephilidae & 0.750 & 0.387 & 0.511 & 31 \\
30 & Sialis & 0.684 & 0.897 & 0.776 & 29 \\
31 & Limnephilus & 0.509 & 1.000 & 0.674 & 29 \\
32 & Micronecta & 0.684 & 0.897 & 0.776 & 29 \\
33 & Bathyomphalus contortus & 0.000 & 0.000 & 0.000 & 28 \\
34 & Pisidium & 1.000 & 0.630 & 0.773 & 27 \\
35 & Sisyra & 0.960 & 0.960 & 0.960 & 25 \\
36 & Caenis rivulorum & 0.000 & 0.000 & 0.000 & 25 \\
37 & Platambus maculatus & 0.952 & 0.870 & 0.909 & 23 \\
38 & Sphaerium & 0.840 & 0.913 & 0.875 & 23 \\
39 & Mystacides longicornis & 1.000 & 0.318 & 0.483 & 22 \\
40 & Mystacides azurea & 0.533 & 0.400 & 0.457 & 20 \\
41 & Erpobdella & 0.000 & 0.000 & 0.000 & 19 \\
42 & Athripsodes cinereus & 1.000 & 0.111 & 0.200 & 18 \\
43 & Heptagenia dalecarlica & 1.000 & 0.625 & 0.769 & 16 \\
44 & Oxyethira & 0.667 & 0.133 & 0.222 & 15 \\
45 & Hydraena & 0.000 & 0.000 & 0.000 & 14 \\
46 & Sialis sordida & 0.600 & 0.214 & 0.316 & 14 \\
47 & Molannodes tinctus & 0.000 & 0.000 & 0.000 & 11 \\
48 & Erythromma najas & 1.000 & 0.400 & 0.571 & 10 \\
49 & Glossiphonia complanata & 1.000 & 0.100 & 0.182 & 10 \\
50 & Neureclipsis bimaculata & 1.000 & 0.222 & 0.364 & 9 \\
51 & Athripsodes aterrimus & 0.000 & 0.000 & 0.000 & 8 \\
52 & Aeshna grandis & 1.000 & 1.000 & 1.000 & 7 \\
53 & Agrypnia & 1.000 & 0.143 & 0.250 & 7 \\
54 & Somatochlora metallica & 1.000 & 1.000 & 1.000 & 6 \\
55 & Oecetis juv. & 0.000 & 0.000 & 0.000 & 6 \\
56 & Platycnemis pennipes & 0.000 & 0.000 & 0.000 & 5 \\
57 & Stylaria lacustris & 0.000 & 0.000 & 0.000 & 5 \\
58 & Ampullaceana balthica & 0.200 & 0.250 & 0.222 & 4 \\
59 & Athripsodes & 0.000 & 0.000 & 0.000 & 4 \\
60 & Myxas glutinosa & 0.000 & 0.000 & 0.000 & 4 \\
61 & Oecetis testacea & 0.667 & 0.667 & 0.667 & 3 \\
62 & Molanna angustata & 0.000 & 0.000 & 0.000 & 3 \\
63 & Tabanidae & 0.000 & 0.000 & 0.000 & 3 \\
64 & Sialis lutaria & 0.000 & 0.000 & 0.000 & 3 \\
65 & Stenochironomus & 0.000 & 0.000 & 0.000 & 2 \\
66 & Mystacides & 0.047 & 1.000 & 0.089 & 2 \\
67 & Leptoceridae juv. & 0.000 & 0.000 & 0.000 & 2 \\
68 & Ceraclea annulicornis & 0.000 & 0.000 & 0.000 & 2 \\
69 & Ischnura elegans & 0.500 & 0.500 & 0.500 & 2 \\
70 & Procladius & 0.000 & 0.000 & 0.000 & 2 \\
71 & Haliplus & 1.000 & 0.500 & 0.667 & 2 \\
72 & Gyraulus albus & 0.000 & 0.000 & 0.000 & 2 \\
73 & Agraylea & 0.000 & 0.000 & 0.000 & 1 \\
74 & Coenagrionidae juv. & 0.000 & 0.000 & 0.000 & 1 \\
75 & Pyralidae & 0.000 & 0.000 & 0.000 & 1 \\
76 & Piscicola geometra & 0.000 & 0.000 & 0.000 & 1 \\
77 & Phryganea bipunctata & 0.000 & 0.000 & 0.000 & 1 \\
78 & Baetis fuscatus & 0.000 & 0.000 & 0.000 & 1 \\
79 & Nemoura avicularis & 0.000 & 0.000 & 0.000 & 1 \\
80 & Ceraclea fulva & 1.000 & 1.000 & 1.000 & 1 \\
81 & Hemiclepsis marginata & 0.000 & 0.000 & 0.000 & 1 \\
82 & Donacia & 0.000 & 0.000 & 0.000 & 1 \\
83 & Cyrnus insolutus & 0.000 & 0.000 & 0.000 & 1 \\
84 & Acroloxus lacustris & 0.000 & 0.000 & 0.000 & 1 \\
\bottomrule
\end{longtable}

\begin{table*}
\caption{Full class-wise standard classification results corresponding to Fig.~\ref{fig:clf_fewshot_classwise}. Index corresponds to the x-axis label in the figure. Support is the number of true examples in the test set.}
\label{tab:classif-fulltaxa}
\centering
\small
\begin{tabular}{llrrrr}
\toprule
Index & Taxon & Precision & Recall & F1-score & Support \\
\midrule
0 & Chironomidae & 0.996 & 0.997 & 0.996 & 2129 \\
1 & Leptophlebia & 0.992 & 0.998 & 0.995 & 1410 \\
2 & Caenis horaria & 0.975 & 0.998 & 0.986 & 822 \\
3 & Asellus aquaticus & 1.000 & 0.997 & 0.998 & 610 \\
4 & Oligochaeta & 0.975 & 0.986 & 0.981 & 591 \\
5 & Kageronia fuscogrisea & 0.995 & 0.993 & 0.994 & 577 \\
6 & Oulimnius tuberculatus & 0.996 & 0.996 & 0.996 & 487 \\
7 & Tinodes waeneri & 0.991 & 1.000 & 0.995 & 218 \\
8 & Ceratopogonidae & 0.994 & 1.000 & 0.997 & 180 \\
9 & Tanypodinae & 0.994 & 0.969 & 0.981 & 159 \\
10 & Micronecta & 1.000 & 0.993 & 0.997 & 145 \\
11 & Caenis luctuosa & 0.989 & 0.870 & 0.926 & 100 \\
12 & Stylaria lacustris & 0.959 & 0.989 & 0.974 & 94 \\
13 & Hydrachnidia & 1.000 & 0.989 & 0.995 & 94 \\
14 & Lepidostoma hirtum & 1.000 & 0.985 & 0.992 & 66 \\
15 & Oulimnius tuberculatus adult & 1.000 & 1.000 & 1.000 & 61 \\
16 & Ephemera vulgata & 1.000 & 1.000 & 1.000 & 57 \\
17 & Helobdella stagnalis & 1.000 & 0.942 & 0.970 & 52 \\
18 & Spirosperma ferox & 0.957 & 0.900 & 0.928 & 50 \\
19 & Ecnomus tenellus & 0.950 & 0.950 & 0.950 & 40 \\
20 & Heptagenia dalecarlica & 1.000 & 0.949 & 0.974 & 39 \\
21 & Cyrnus trimaculatus & 0.968 & 0.833 & 0.896 & 36 \\
22 & Erpobdella octoculata & 0.943 & 1.000 & 0.971 & 33 \\
23 & Psychomyia pusilla & 1.000 & 1.000 & 1.000 & 33 \\
24 & Gyraulus & 0.938 & 1.000 & 0.968 & 30 \\
25 & Cloeon dipterum & 0.929 & 0.867 & 0.897 & 30 \\
26 & Mystacides azurea & 1.000 & 0.929 & 0.963 & 28 \\
27 & Limnephilus & 0.852 & 0.885 & 0.868 & 26 \\
28 & Polycentropus flavomaculatus & 0.897 & 1.000 & 0.945 & 26 \\
29 & Turbellaria & 0.947 & 0.750 & 0.837 & 24 \\
30 & Cloeon & 0.913 & 0.913 & 0.913 & 23 \\
31 & Cyrnus flavidus & 0.957 & 1.000 & 0.978 & 22 \\
32 & Hydroptila & 0.875 & 1.000 & 0.933 & 21 \\
33 & Centroptilum luteolum & 0.933 & 0.778 & 0.848 & 18 \\
34 & Limnephilidae & 0.800 & 0.706 & 0.750 & 17 \\
35 & Pisidium & 1.000 & 0.882 & 0.938 & 17 \\
36 & Pedicia & 1.000 & 1.000 & 1.000 & 13 \\
37 & Orthotrichia & 1.000 & 0.917 & 0.957 & 12 \\
38 & Sialis & 1.000 & 1.000 & 1.000 & 12 \\
39 & Mystacides & 0.778 & 0.636 & 0.700 & 11 \\
40 & Caenis rivulorum & 0.900 & 0.900 & 0.900 & 10 \\
41 & Sisyra & 1.000 & 0.900 & 0.947 & 10 \\
\bottomrule
\end{tabular}
\end{table*}

\begin{table*}
\caption{Full class-wise few-shot classification results corresponding to Fig.~\ref{fig:clf_fewshot_classwise}.  Index corresponds to the x-axis label in the figure. Support is the number of true examples after aggregating all test sets of the five cross-validation folds.}
\label{tab:fewshot-fulltaxa}
\centering
\small
\begin{tabular}{llrrrr}
\toprule
Index & Taxon & Precision & Recall & F1-score & Support \\
\midrule
42 & Oxyethira & 0.894 & 0.955 & 0.923 & 44 \\
43 & Sphaerium & 0.951 & 0.907 & 0.929 & 43 \\
44 & Erpobdella & 0.953 & 0.976 & 0.965 & 42 \\
45 & Bathyomphalus contortus & 0.857 & 1.000 & 0.923 & 42 \\
46 & Athripsodes cinereus & 0.660 & 0.756 & 0.705 & 41 \\
47 & Agrypnia & 0.778 & 0.757 & 0.767 & 37 \\
48 & Platambus maculatus & 0.914 & 0.865 & 0.889 & 37 \\
49 & Mystacides longicornis & 0.921 & 0.972 & 0.946 & 36 \\
50 & Nematoda & 0.921 & 0.972 & 0.946 & 36 \\
51 & Sialis sordida & 0.838 & 0.861 & 0.849 & 36 \\
52 & Haliplus & 0.914 & 0.914 & 0.914 & 35 \\
53 & Nemoura & 0.675 & 0.818 & 0.740 & 33 \\
54 & Glossiphonia complanata & 0.958 & 0.920 & 0.939 & 25 \\
55 & Somatochlora metallica & 0.767 & 1.000 & 0.868 & 23 \\
56 & Nemoura avicularis & 0.733 & 0.478 & 0.579 & 23 \\
57 & Ampullaceana balthica & 0.857 & 0.900 & 0.878 & 20 \\
58 & Cyrnus juv. & 0.947 & 0.947 & 0.947 & 19 \\
59 & Molannodes tinctus & 0.625 & 0.588 & 0.606 & 17 \\
60 & Athripsodes aterrimus & 0.600 & 0.529 & 0.562 & 17 \\
61 & Sialis lutaria & 0.647 & 0.647 & 0.647 & 17 \\
62 & Hydraena & 1.000 & 1.000 & 1.000 & 17 \\
63 & Ischnura elegans & 0.733 & 0.647 & 0.688 & 17 \\
64 & Neureclipsis bimaculata & 0.941 & 1.000 & 0.970 & 16 \\
65 & Mystacides juv. & 0.700 & 0.875 & 0.778 & 16 \\
66 & Oecetis testacea & 0.882 & 0.938 & 0.909 & 16 \\
67 & Erythromma najas & 0.750 & 0.643 & 0.692 & 14 \\
68 & Molanna angustata & 0.429 & 0.250 & 0.316 & 12 \\
69 & Ripistes parasita & 0.923 & 1.000 & 0.960 & 12 \\
70 & Athripsodes juv. & 0.571 & 0.667 & 0.615 & 12 \\
71 & Oecetis juv. & 0.846 & 0.917 & 0.880 & 12 \\
72 & Aeshna grandis & 0.800 & 1.000 & 0.889 & 12 \\
73 & Ceraclea annulicornis & 0.800 & 0.727 & 0.762 & 11 \\
74 & Athripsodes & 0.667 & 0.364 & 0.471 & 11 \\
75 & Gyraulus albus & 1.000 & 0.300 & 0.462 & 10 \\
76 & Platycnemis pennipes & 0.714 & 0.556 & 0.625 & 9 \\
77 & Baetis fuscatus & 1.000 & 0.778 & 0.875 & 9 \\
78 & Tabanidae & 0.875 & 0.875 & 0.875 & 8 \\
79 & Heptagenia sulphurea & 1.000 & 1.000 & 1.000 & 7 \\
80 & Piscicola geometra & 0.667 & 0.333 & 0.444 & 6 \\
81 & Myxas glutinosa & 0.714 & 0.833 & 0.769 & 6 \\
82 & Hygrotus & 1.000 & 0.667 & 0.800 & 6 \\
83 & Holocentropus picicornis & 1.000 & 0.833 & 0.909 & 6 \\
84 & Nematomorpha & 0.750 & 0.600 & 0.667 & 5 \\
85 & Ceraclea nigronervosa & 0.667 & 0.800 & 0.727 & 5 \\
86 & Ceraclea fulva & 0.600 & 0.600 & 0.600 & 5 \\
87 & Stenochironomus & 0.833 & 1.000 & 0.909 & 5 \\
88 & Leptoceridae juv. & 1.000 & 0.400 & 0.571 & 5 \\
\bottomrule
\end{tabular}
\end{table*}

\clearpage

\subsection{Biomass estimation}

\begin{figure}
    \centering
    \includegraphics[width=0.7\linewidth]{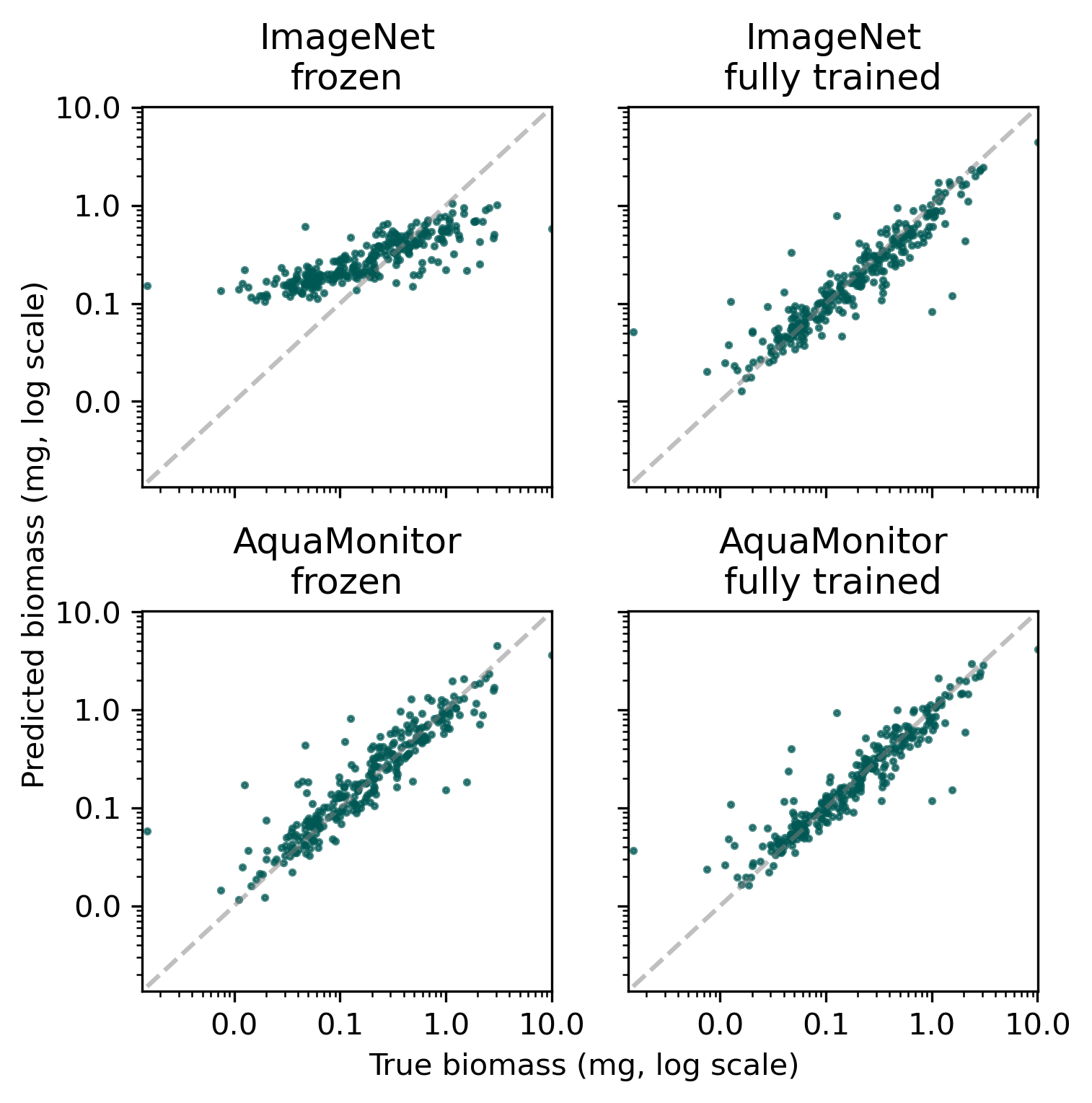}
    \caption{Scatter plots of biomass estimation regression task with different training approaches. Diagonal line represents perfect prediction.}
    \label{fig:biomass-scatter}
\end{figure}

Fig.~\ref{fig:biomass-scatter} shows scatter plots for the biomass estimation task reported in Table~7 in the main paper. ImageNet denotes that the model used only ImageNet weights for pretraining, AquaMonitor denotes the weights of the best performing Swin-T model from the standard classification task were used. Frozen models have all layers except the final feed-forward layer frozen, while fully trained models have all parameters trainable. The figure further illustrates how the representations learned from the classification task transfer to the regression task better than just ImageNet weights.
Full biomass evaluation table with bootstrapped 2-sigma confidence intervals corresponding to Table~7 from the main paper can be seen in Table~\ref{tab:biomass_fulltable}.

\begin{table}
\centering
\caption{Biomass evaluation results corresponding to Table~7 in the main paper, with bootstrapped 2-sigma confidence intervals.}
\label{tab:biomass_fulltable}
\begin{tabular}{cllll}
\toprule
Dataset & Frozen & MdAPE & MAE & MAPE \\
\midrule
ImageNet &  \checkmark & 0.6856 ($\pm$0.125) & 0.2512 ($\pm$0.078) & 1.8584 ($\pm$0.731) \\
ImageNet &  & 0.1957 ($\pm$0.021) & 0.1197 ($\pm$0.044) & 0.4568 ($\pm$0.248) \\
AquaMonitor & \checkmark & 0.2410 ($\pm$0.039) & 0.1436 ($\pm$0.051) & 0.5817 ($\pm$0.277) \\
AquaMonitor &  & 0.1733 ($\pm$0.029) & 0.1128 ($\pm$0.046) & 0.4306 ($\pm$0.186) \\
\bottomrule
\end{tabular}
\end{table}

\subsection{Out-of-distribution detection}

Table~\ref{tab:ood-appendix} gives the results of out-of-distribution detection for 72 outlier specimens from 24 classes, corresponding to Fig.~5 in the main paper. OOD detection is performed with the same classifier model that was trained for the monitoring benchmark. Entropy and energy~\cite{liu2020Energybased} scores are calculated from the softmax output for each specimen, while MaxLogit~\cite{hendrycks2022Scaling} is calculated from the raw logit output. Entropy is the standard entropy score for the output distribution.
It can be seen that although MaxLogit performs overall the best for all models, the best OOD detection performance is gained with the multiview model, using the entropy ranking metric.

\begin{table}[htb]
\centering
\small
\caption{Out-of-distribution detection AUROC results for the monitoring dataset, with different OOD ranking metrics. Confidence intervals are bootstrapped 2-sigma intervals with 1000 repetitions.}
\label{tab:ood-appendix}
\begin{tabular}{lrrr}
\toprule
Model & Energy & Entropy & MaxLogit \\
\midrule
MobileNetV3 & 0.7123 ($\pm$0.052) & 0.7130 ($\pm$0.053) & 0.6918 ($\pm$0.058) \\
ResNet-50 & 0.7452 ($\pm$0.047) & 0.7473 ($\pm$0.048) & 0.7603 ($\pm$0.048) \\
ResNet-101 & 0.7351 ($\pm$0.041) & 0.7375 ($\pm$0.043) & 0.7169 ($\pm$0.060) \\
EfficientNet-B0 & 0.7472 ($\pm$0.051) & 0.7471 ($\pm$0.050) & 0.8008 ($\pm$0.044) \\
EfficientNet-B4 & 0.7564 ($\pm$0.042) & 0.7570 ($\pm$0.042) & 0.7288 ($\pm$0.063) \\
Swin-T & 0.7742 ($\pm$0.031) & 0.7746 ($\pm$0.032) & 0.7892 ($\pm$0.034) \\
Swin-T (Multiview) & 0.8230 ($\pm$0.032) & 0.8235 ($\pm$0.030) & 0.7972 ($\pm$0.040) \\
Swin-B & 0.7507 ($\pm$0.044) & 0.7516 ($\pm$0.043) & 0.7249 ($\pm$0.049) \\
ViT-B/16 & 0.7207 ($\pm$0.045) & 0.7266 ($\pm$0.046) & 0.7491 ($\pm$0.056) \\
ViT-B/16 (BioCLIP) & 0.7196 ($\pm$0.046) & 0.7215 ($\pm$0.046) & 0.7500 ($\pm$0.052) \\
ViT-B/16 (BioCLIP-FT) & 0.7727 ($\pm$0.039) & 0.7733 ($\pm$0.041) & 0.7706 ($\pm$0.047) \\
ViT-L/14 & 0.7706 ($\pm$0.040) & 0.7740 ($\pm$0.039) & 0.7778 ($\pm$0.046) \\
Ensemble & 0.7857 ($\pm$0.033) & 0.7864 ($\pm$0.033) & 0.8046 ($\pm$0.039) \\
\bottomrule
\end{tabular}
\end{table}

%% file: full_table.tex
\begin{table}[htb]
\caption{Full results on Monitoring benchmark. BioCLIP refers to fully trained ViT-B/16 starting from BioCLIP weights. BioCLIP-FT refers to a ViT-B/16 using BioCLIP weights, but only the last two transformer blocks being trainable.}
\label{tab:monitor-fulltable}
\centering
\small
\begin{tabular}{lllll}
\toprule
Model & Accuracy & Top-3 & Top-5 & F1 weighted \\
\midrule
MobileNetV3 & 0.7514 ($\pm$0.006) & 0.9006 ($\pm$0.004) & 0.9405 ($\pm$0.003) & 0.7132 ($\pm$0.008) \\
ResNet-50 & 0.8508 ($\pm$0.005) & 0.9415 ($\pm$0.004) & 0.9630 ($\pm$0.003) & 0.8257 ($\pm$0.007) \\
ResNet-101 & 0.8573 ($\pm$0.005) & 0.9392 ($\pm$0.003) & 0.9594 ($\pm$0.003) & 0.8343 ($\pm$0.006) \\
EfficientNet-B0 & 0.8562 ($\pm$0.005) & 0.9518 ($\pm$0.003) & 0.9719 ($\pm$0.002) & 0.8361 ($\pm$0.006) \\
EfficientNet-B4 & 0.8669 ($\pm$0.005) & 0.9463 ($\pm$0.003) & 0.9654 ($\pm$0.003) & 0.8430 ($\pm$0.006) \\
Swin-T & 0.8696 ($\pm$0.005) & 0.9701 ($\pm$0.003) & 0.9848 ($\pm$0.002) & 0.8497 ($\pm$0.006) \\
Swin-T (Multiview) & 0.8791 ($\pm$0.005) & 0.9652 ($\pm$0.003) & 0.9805 ($\pm$0.002) & 0.8580 ($\pm$0.006) \\
Swin-B & 0.8581 ($\pm$0.005) & 0.9632 ($\pm$0.003) & 0.9802 ($\pm$0.002) & 0.8383 ($\pm$0.006) \\
ViT-B/16 & 0.8112 ($\pm$0.006) & 0.9510 ($\pm$0.003) & 0.9692 ($\pm$0.003) & 0.7995 ($\pm$0.006) \\
ViT-B/16 (BioCLIP) & 0.8166 ($\pm$0.006) & 0.9381 ($\pm$0.004) & 0.9596 ($\pm$0.003) & 0.7904 ($\pm$0.007) \\
ViT-B/16 (BioCLIP-FT) & 0.8352 ($\pm$0.006) & 0.9429 ($\pm$0.003) & 0.9682 ($\pm$0.003) & 0.8086 ($\pm$0.007) \\
ViT-L/14 & 0.8538 ($\pm$0.005) & 0.9582 ($\pm$0.003) & 0.9749 ($\pm$0.002) & 0.8375 ($\pm$0.006) \\
Ensemble & 0.8818 ($\pm$0.005) & 0.9680 ($\pm$0.003) & 0.9839 ($\pm$0.002) & 0.8591 ($\pm$0.006) \\
\end{tabular}

\begin{tabular}{lllll}
\toprule
Model & F1 macro & Precision macro & Precision weighted & Recall macro \\
\midrule
MobileNetV3 & 0.2281 ($\pm$0.023) & 0.3440 ($\pm$0.039) & 0.7811 ($\pm$0.009) & 0.2341 ($\pm$0.025) \\
ResNet-50 & 0.3273 ($\pm$0.025) & 0.4690 ($\pm$0.039) & 0.8659 ($\pm$0.006) & 0.3249 ($\pm$0.027) \\
ResNet-101 & 0.3224 ($\pm$0.024) & 0.4084 ($\pm$0.031) & 0.8599 ($\pm$0.006) & 0.3245 ($\pm$0.026) \\
EfficientNet-B0 & 0.3055 ($\pm$0.025) & 0.4073 ($\pm$0.036) & 0.8629 ($\pm$0.006) & 0.3137 ($\pm$0.029) \\
EfficientNet-B4 & 0.3152 ($\pm$0.025) & 0.4357 ($\pm$0.037) & 0.8749 ($\pm$0.006) & 0.3092 ($\pm$0.027) \\
Swin-T & 0.3610 ($\pm$0.030) & 0.4856 ($\pm$0.038) & 0.8791 ($\pm$0.005) & 0.3428 ($\pm$0.030) \\
Swin-T (Multiview) & 0.3378 ($\pm$0.024) & 0.4440 ($\pm$0.032) & 0.8816 ($\pm$0.006) & 0.3316 ($\pm$0.026) \\
Swin-B & 0.3354 ($\pm$0.026) & 0.4230 ($\pm$0.033) & 0.8687 ($\pm$0.005) & 0.3348 ($\pm$0.028) \\
ViT-B/16 & 0.2923 ($\pm$0.024) & 0.3946 ($\pm$0.032) & 0.8417 ($\pm$0.006) & 0.2705 ($\pm$0.025) \\
ViT-B/16 (BioCLIP) & 0.2579 ($\pm$0.019) & 0.3739 ($\pm$0.030) & 0.8357 ($\pm$0.007) & 0.2498 ($\pm$0.020) \\
ViT-B/16 (BioCLIP-FT) & 0.3256 ($\pm$0.027) & 0.4523 ($\pm$0.040) & 0.8519 ($\pm$0.006) & 0.3238 ($\pm$0.028) \\
ViT-L/14 & 0.3146 ($\pm$0.024) & 0.4306 ($\pm$0.036) & 0.8607 ($\pm$0.006) & 0.2940 ($\pm$0.024) \\
Ensemble & 0.3673 ($\pm$0.029) & 0.4915 ($\pm$0.038) & 0.8834 ($\pm$0.005) & 0.3596 ($\pm$0.030) \\
\bottomrule
\end{tabular}
\end{table}

\begin{table}[htb]
\caption{Full results on classification benchmark.}
\label{tab:classif-fulltable}
\centering
\small
\begin{tabular}{lllll}
\toprule
Model & Accuracy & Top-3 & Top-5 & F1 weighted \\
\midrule
MobileNetV3 & 0.9693 ($\pm$0.004) & 0.9949 ($\pm$0.002) & 0.9972 ($\pm$0.001) & 0.9683 ($\pm$0.004) \\
ResNet-50 & 0.9809 ($\pm$0.003) & 0.9967 ($\pm$0.001) & 0.9980 ($\pm$0.001) & 0.9804 ($\pm$0.003) \\
EfficientNet-B0 & 0.9832 ($\pm$0.003) & 0.9968 ($\pm$0.001) & 0.9980 ($\pm$0.001) & 0.9826 ($\pm$0.003) \\
EfficientNet-B4 & 0.9848 ($\pm$0.003) & 0.9985 ($\pm$0.001) & 0.9989 ($\pm$0.001) & 0.9846 ($\pm$0.003) \\
Swin-T & 0.9879 ($\pm$0.002) & 0.9982 ($\pm$0.001) & 0.9993 ($\pm$0.001) & 0.9877 ($\pm$0.003) \\
Swin-T (Multiview) & 0.9850 ($\pm$0.003) & 0.9974 ($\pm$0.001) & 0.9985 ($\pm$0.001) & 0.9847 ($\pm$0.003) \\
Ensemble & 0.9879 ($\pm$0.002) & 0.9982 ($\pm$0.001) & 0.9988 ($\pm$0.001) & 0.9877 ($\pm$0.002) \\
\end{tabular}

\begin{tabular}{lllll}
\toprule
Model & F1 macro & Precision macro & Precision weighted & Recall macro \\
\midrule
MobileNetV3 & 0.9039 ($\pm$0.017) & 0.9430 ($\pm$0.015) & 0.9692 ($\pm$0.004) & 0.8770 ($\pm$0.020) \\
ResNet-50 & 0.9258 ($\pm$0.016) & 0.9554 ($\pm$0.014) & 0.9809 ($\pm$0.003) & 0.9035 ($\pm$0.019) \\
EfficientNet-B0 & 0.9329 ($\pm$0.016) & 0.9557 ($\pm$0.014) & 0.9831 ($\pm$0.003) & 0.9173 ($\pm$0.017) \\
EfficientNet-B4 & 0.9443 ($\pm$0.014) & 0.9652 ($\pm$0.010) & 0.9851 ($\pm$0.003) & 0.9288 ($\pm$0.017) \\
Swin-T & 0.9455 ($\pm$0.015) & 0.9605 ($\pm$0.013) & 0.9880 ($\pm$0.002) & 0.9343 ($\pm$0.017) \\
Swin-T (Multiview) & 0.9375 ($\pm$0.016) & 0.9548 ($\pm$0.014) & 0.9851 ($\pm$0.003) & 0.9243 ($\pm$0.018) \\
Ensemble & 0.9484 ($\pm$0.014) & 0.9617 ($\pm$0.013) & 0.9879 ($\pm$0.002) & 0.9380 ($\pm$0.016) \\
\bottomrule
\end{tabular}
\end{table}

\begin{table}[htb]
\caption{Full results on few-shot benchmark.}
\label{tab:fewshot-fulltable}
\centering
\small
\begin{tabular}{lllll}
\toprule
Model & Accuracy & Top-3 & Top-5 & F1 weighted \\
\midrule
MobileNetV3 & 0.7380 ($\pm$0.029) & 0.8947 ($\pm$0.020) & 0.9104 ($\pm$0.019) & 0.7307 ($\pm$0.031) \\
ResNet-50 & 0.7783 ($\pm$0.029) & 0.9239 ($\pm$0.018) & 0.9362 ($\pm$0.017) & 0.7749 ($\pm$0.030) \\
EfficientNet-B0 & 0.8063 ($\pm$0.027) & 0.9149 ($\pm$0.019) & 0.9272 ($\pm$0.017) & 0.7949 ($\pm$0.029) \\
EfficientNet-B4 & 0.8275 ($\pm$0.024) & 0.9306 ($\pm$0.017) & 0.9418 ($\pm$0.015) & 0.8213 ($\pm$0.026) \\
Swin-T & 0.8287 ($\pm$0.025) & 0.9250 ($\pm$0.018) & 0.9362 ($\pm$0.016) & 0.8216 ($\pm$0.027) \\
\midrule
CLIP/BioCLIP & 0.7592 ($\pm$0.029) & 0.9104 ($\pm$0.019) & 0.9373 ($\pm$0.017) & 0.7538 ($\pm$0.030) \\
CLIP/OpenAI & 0.7234 ($\pm$0.031) & 0.8779 ($\pm$0.023) & 0.9183 ($\pm$0.019) & 0.7108 ($\pm$0.032) \\
CLIP/OpenCLIP & 0.7100 ($\pm$0.030) & 0.8667 ($\pm$0.022) & 0.9071 ($\pm$0.019) & 0.7028 ($\pm$0.031) \\
DINO & 0.7581 ($\pm$0.028) & 0.8891 ($\pm$0.021) & 0.9127 ($\pm$0.019) & 0.7527 ($\pm$0.029) \\
DINOv2 & 0.7548 ($\pm$0.030) & 0.9037 ($\pm$0.020) & 0.9328 ($\pm$0.017) & 0.7482 ($\pm$0.031) \\
SigLIP & 0.7335 ($\pm$0.029) & 0.9026 ($\pm$0.019) & 0.9373 ($\pm$0.016) & 0.7264 ($\pm$0.030) \\
SigLIP2 & 0.7279 ($\pm$0.030) & 0.9003 ($\pm$0.021) & 0.9317 ($\pm$0.017) & 0.7208 ($\pm$0.031) \\
\end{tabular}

\begin{tabular}{lllll}
\toprule
Model & F1 macro & Precision macro & Precision weighted & Recall macro \\
\midrule
MobileNetV3 & 0.6983 ($\pm$0.037) & 0.7036 ($\pm$0.041) & 0.7295 ($\pm$0.032) & 0.7008 ($\pm$0.037) \\
ResNet-50 & 0.7437 ($\pm$0.037) & 0.7705 ($\pm$0.037) & 0.7819 ($\pm$0.030) & 0.7382 ($\pm$0.037) \\
EfficientNet-B0 & 0.7449 ($\pm$0.032) & 0.7745 ($\pm$0.033) & 0.7985 ($\pm$0.030) & 0.7411 ($\pm$0.031) \\
EfficientNet-B4 & 0.7785 ($\pm$0.035) & 0.8246 ($\pm$0.042) & 0.8350 ($\pm$0.026) & 0.7695 ($\pm$0.035) \\
Swin-T & 0.7805 ($\pm$0.037) & 0.8143 ($\pm$0.040) & 0.8291 ($\pm$0.026) & 0.7737 ($\pm$0.036) \\
\midrule
CLIP/BioCLIP & 0.7202 ($\pm$0.032) & 0.7478 ($\pm$0.036) & 0.7641 ($\pm$0.031) & 0.7175 ($\pm$0.031) \\
CLIP/OpenAI & 0.6505 ($\pm$0.039) & 0.7278 ($\pm$0.054) & 0.7325 ($\pm$0.034) & 0.6395 ($\pm$0.036) \\
CLIP/OpenCLIP & 0.6474 ($\pm$0.035) & 0.6801 ($\pm$0.045) & 0.7151 ($\pm$0.033) & 0.6455 ($\pm$0.034) \\
DINO & 0.7093 ($\pm$0.034) & 0.7583 ($\pm$0.044) & 0.7713 ($\pm$0.031) & 0.7044 ($\pm$0.032) \\
DINOv2 & 0.6960 ($\pm$0.038) & 0.7588 ($\pm$0.040) & 0.7670 ($\pm$0.029) & 0.6802 ($\pm$0.037) \\
SigLIP & 0.6548 ($\pm$0.036) & 0.7126 ($\pm$0.046) & 0.7463 ($\pm$0.031) & 0.6495 ($\pm$0.034) \\
SigLIP2 & 0.6635 ($\pm$0.036) & 0.7204 ($\pm$0.041) & 0.7386 ($\pm$0.031) & 0.6541 ($\pm$0.034) \\
\bottomrule
\end{tabular}

\end{table}